\documentclass[a4paper,fleqn]{cas-dc}

\usepackage[numbers]{natbib}

\usepackage{algorithm}
\usepackage{algorithmic}
\newtheorem{theorem}{Theorem}
\newtheorem{lemma}{Lemma}
\newproof{pf}{Proof}

\usepackage{adjustbox}
\usepackage{threeparttable}
\usepackage{multirow} % Required for multirows 
\usepackage{soul, color, xcolor}

\def\tsc#1{\csdef{#1}{\textsc{\lowercase{#1}}\xspace}}
\tsc{WGM}
\tsc{QE}
\tsc{EP}
\tsc{PMS}
\tsc{BEC}
\tsc{DE}
\begin{document}
\let\WriteBookmarks\relax
\def\floatpagepagefraction{1}
\def\textpagefraction{.001}
\shorttitle{Bit-adaptive Spiking Neural Networks}
\shortauthors{Yao et~al.}

\title [mode = title]{Towards Efficient and Accurate Spiking Neural Networks via Adaptive Bit Allocation}                      
% \tnotemark[1,2]

% \tnotetext[1]{This document is the results of the research
%    project funded by the National Science Foundation.}

% \tnotetext[2]{The second title footnote which is a longer text matter
%    to fill through the whole text width and overflow into
%    another line in the footnotes area of the first page.}

\author[1,2]{Yao Xingting}[style=chinese,orcid=0009-0000-1872-1170]
\ead{yaoxingting2020@ia.ac.cn}
\cormark[1]
\credit{Writing – original draft, Writing – review \& editing, Conceptualization, Investigation, Data curation, Formal analysis, Software, Project administration, Resources}
\affiliation[1]{organization={The Key Laboratory of Cognition and Decision Intelligence for Complex Systems, Institute of Automation, Chinese Academy of Sciences},
                addressline={Zhongguan Cun East Road No.95}, 
                city={Beijing},
                postcode={100190}, 
                country={China}}

\affiliation[2]{organization={School of Future Technology, University of Chinese Academy of Sciences},        
                addressline={Huaibei Zhuang No.380}, 
                city={Beijing},
                postcode={101408}, 
                country={China}}

\author[1]{Hu Qinghao}[style=chinese,orcid=0000-0003-0422-5509]
\cormark[1]
\ead{huqinghao2014@ia.ac.cn}
\credit{Writing – review \& editing, Methodology, Formal analysis,  Supervision}

\author[3]{Zhou Fei}[style=chinese,orcid=0009-0006-8357-860X]
\ead{1242643224@qq.com}
\credit{Writing – review \& editing, Funding acquisition}
\affiliation[3]{organization={China Electric Power Research Institute Co., Ltd},
                addressline={Future Science Park Binhe Avenue No.18}, 
                city={Beijing},
%               citysep={}, % Uncomment if no comma needed between city and postcode
                postcode={102211}, 
                country={China}}

\author[1,2]{Liu Tielong}[style=chinese]
\ead{liutielong2022@ia.ac.cn}
\credit{Writing – review \& editing, Formal analysis,}

\author[1]{Li Gang}[style=chinese,orcid=0000-0001-7835-4739]
\ead{gang.li@ia.ac.cn}
\credit{Writing – review \& editing}

\author[1]{Wang Peisong}[style=chinese]
\ead{wangpeisong2013@ia.ac.cn}
\credit{Writing – review \& editing}

\author[1,2]{Cheng Jian}[style=chinese,orcid=0000-0003-1289-2758]
\cormark[2]
\ead{jian.cheng@ia.ac.cn}
\credit{Writing – review \& editing, Funding acquisition}

\cortext[cor1]{Equal contribution}
\cortext[cor2]{Corresponding author}
% \cortext[cor2]{Principal corresponding author}
% % \fntext[fn1]{This is the first author footnote, but is common to third
% %   author as well.}
% % \fntext[fn2]{Another author footnote, this is a very long footnote and
% %   it should be a really long footnote. But this footnote is not yet
% %   sufficiently long enough to make two lines of footnote text.}

% % \nonumnote{This note has no numbers. In this work we demonstrate $a_b$
% %   the formation Y\_1 of a new type of polariton on the interface
% %   between a cuprous oxide slab and a polystyrene micro-sphere placed
% %   on the slab.
% %   }

\begin{abstract}
Multi-bit spiking neural networks (SNNs) have recently become a heated research spot, pursuing energy-efficient and high-accurate AI. 
However, with more bits involved, the associated memory and computation demands escalate to the point where the performance improvements become disproportionate. Based on the insight that different layers demonstrate different importance and extra bits could be wasted and interfering, this paper presents an adaptive bit allocation strategy for direct-trained SNNs, achieving fine-grained layer-wise allocation of memory and computation resources. Thus, SNN's efficiency and accuracy can be improved. 
Specifically, we parametrize the temporal lengths and the bit widths of weights and spikes, and make them learnable and controllable through gradients. To address the challenges caused by changeable bit widths and temporal lengths, we propose 
the refined spiking neuron, which can handle different temporal lengths, enable the derivation of gradients for temporal lengths, and suit spike quantization better. In addition, we theoretically formulate the step-size mismatch problem of learnable bit widths, which may incur severe quantization errors to SNN, and accordingly propose the step-size renewal mechanism to alleviate this issue.  Experiments on various datasets, including the static CIFAR and ImageNet datasets and the dynamic CIFAR-DVS and DVS-GESTURE datasets, demonstrate that our methods can reduce the overall memory and computation cost while achieving higher accuracy. Particularly, our SEWResNet-34 can achieve a 2.69\% accuracy gain and 4.16$\times$ lower bit budgets over the advanced baseline work on ImageNet. This work is open-sourced at \href{https://github.com/Ikarosy/Towards-Efficient-and-Accurate-Spiking-Neural-Networks-via-Adaptive-Bit-Allocation}{this link}.
\end{abstract}

% \includegraphics{figs/cas-grabs.pdf}

% \begin{highlights}
% \item Research highlights item 1
% \item Research highlights item 2
% \item Research highlights item 3
% \end{highlights}


\maketitle

\section{Introduction}
\label{sec:intro}

\begin{figure*}[t]
  \centering
  \includegraphics[width= 0.7\textwidth]{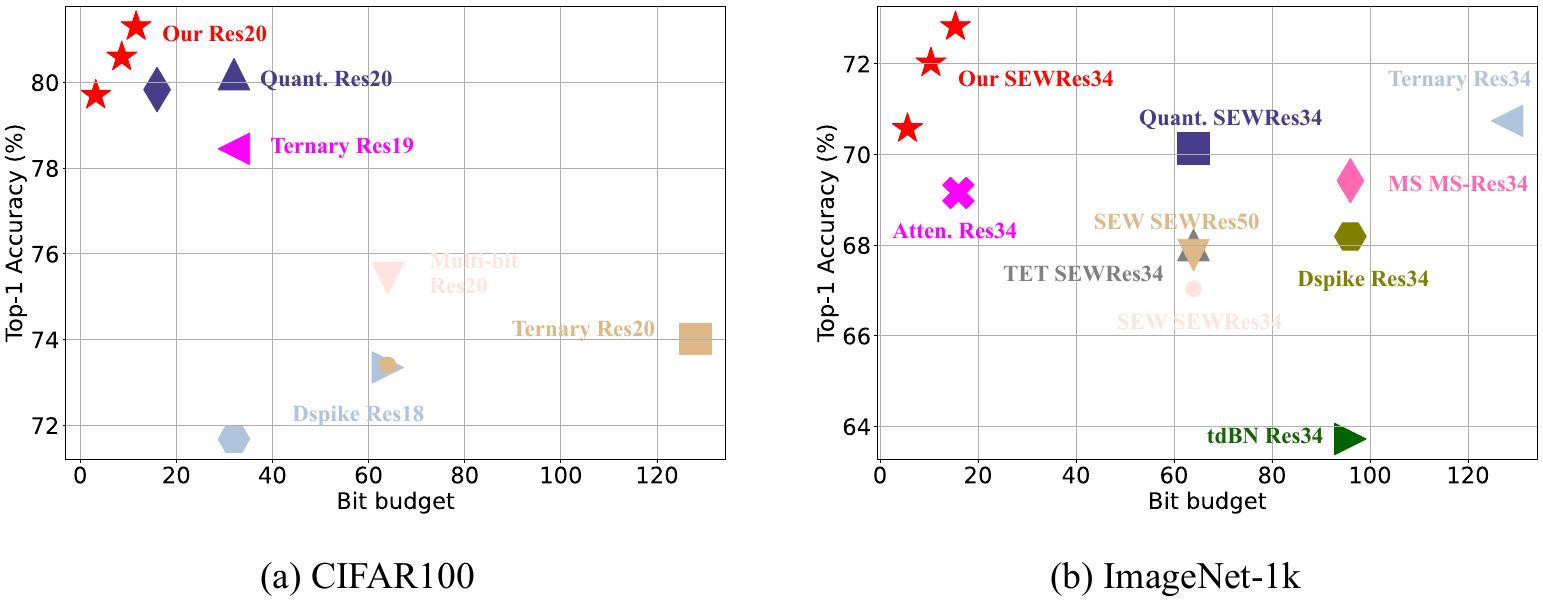}
  \caption{Comparisons with other advanced direct-trained SNNs using ResNet-based architectures on CIFAR100 and ImageNet-1k. Our models maintain the same level of model size as the baselines.}
  \label{fig:intro}
\end{figure*}

Spiking neural networks (SNNs) are considered the third generation of neural networks due to their emulation of the human brain's information-processing mechanism \citep{maass1997networks}. 
The essential uniqueness of SNN is its activation unit, i.e., spiking neuron, which transfers floating-point  values into binary spikes, thus making the matrix operations, e.g., convolution, conducted multiplication-free \citep{fang2021deep}.  Similar to binary neural networks (BNN), binary feature maps (spikes) would reduce spatial representation capacity, making processing information-dense data challenging and resulting in lower model accuracy \citep{zhu2019binary, meng2022training,zhou2022spikformer}.

In this context, multi-bit spiking neural networks have recently arisen as a promising approach to improving such inadequate informational scope \citep{guo2024ternary, xing2024spikellm,xing2024spikelm,xiao2024multi}. As the name suggests, multi-bit SNNs introduce more bits to represent the original binary spike and devote more addition operations to process the matrix multiplication. Prior arts successfully incorporated this mechanism to improve accuracy, while most neglected the actual memory and computation overhead as pointed out by  \citep{shen2024conventional}.
For instance, a layer with less representation ability but assigned with higher bit widths would incur memory and computation waste and interfere with  the model inference \citep{dong2019hawq,Chen_2021_ICCV}. 
It necessarily comes to the 
question: \emph{Can we enhance SNN's performance while keeping memory and computation costs at minimal levels?}

\begin{figure*}[t]
  \centering
  \includegraphics[width= 1.\textwidth]{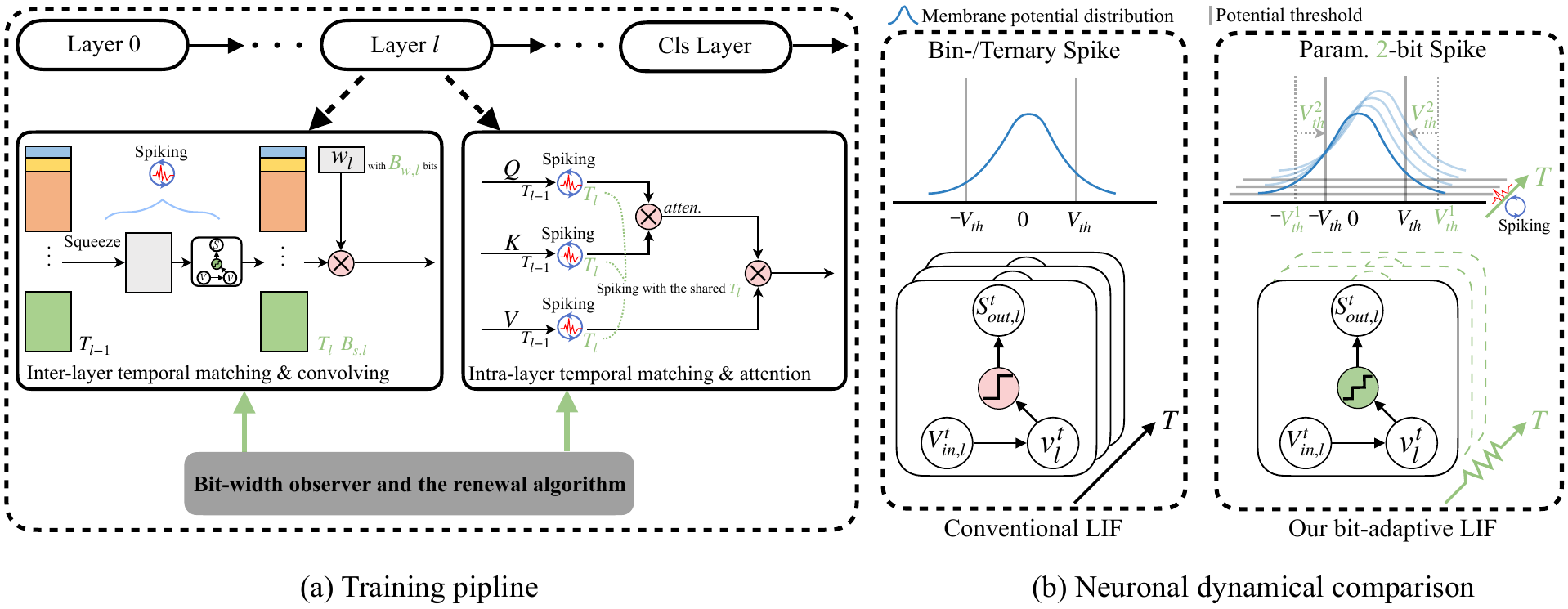}
  \caption{Overview of the  proposed  bit allocation method. Green notations denote the parametrized constants. \textbf{(a)} Adaptive-bit-allocation training pipline, where the bit widths $T_l$, $B_{s,l}$, and $B_{w,l}$ are made learnable and adaptive. $T_l$ is matched inter-layer and intra-layer to ensure fluent dataflow. The step-size renewal mechanism is also proposed to alleviate the step-size mismatch issue. \textbf{(b)} depicts the neuron formulation (bottom) and the potential division (top) of the previous spiking neuron (left) and our bit-adaptive spiking neuron (right). The temporal length $T_l$ is layer-wise learnable and a shift $V^2_{th,l}$ is added to the potential threshold $V^1_{th,l}$. }
  \label{fig:overview}
\end{figure*}
In this paper, we investigate the adaptive bit allocation of quantized SNNs as the answer. Specifically, we parametrize the three fundamental elements of memory and computation: temporal lengths, bit widths of weights, and bit widths of spikes, making them learnable and able to be optimized end-to-end. 
To regulate the allocation process of temporal lengths and bit widths during training and ensure they converge to the targets correctly, we explicitly set temporal length and bit width bounds and implicitly design the regulating loss. 
By implementing these strategies, we can effectively assign different temporal lengths and bit widths to different layers, enabling a fine-grained 
allocation of memory and computation resources for SNN. Thus, the accuracy and efficiency of the inference can be improved.
Moreover, unlike the memory and computation measurement of most prior arts \citep{zhou2022spikformer, guo2024ternary, xing2024spikelm, yao2023attention}, we 
incorporate the pragmatic concept "Bit Budget" and the associated S-ACE, which are proposed by  \citep{shen2024conventional}, to estimate the accurate model memory and computation overheads.

With the above quantization and parametrization methods applied, new challenges concerning the unique temporal dimension of SNNs arise. 
Firstly, the length of the temporal dimension is a hyperparameter and initially does not get involved in the forward pass, thus 
failing to get gradients for optimization. 
Secondly, previous multi-bit spiking neurons are designed and tuned in a relatively arbitrary and empirical manner, showing less friendliness to quantization. Thirdly, the potential step-size mismatch issue of learnable bit widths may become more severe. Because once a wrong step size is applied, it will be consecutively used along the temporal dimension.
As illustrated in   Figure~\ref{fig:overview}, we first refine the spiking neuron, which addresses inter-layer and intra-layer temporal mismatches arising from variable temporal lengths while minimizing quantization errors through improved formulations and potential division operations. 
Then, we formulate the quantization step-size mismatch issue, where SNN would suffer statistically more quantization errors that accumulate along the temporal dimension.
Accordingly,
we propose the step-size renewal mechanism, which consists of a bit-width observer and the renewal algorithm as depicted in  Figure~\ref{fig:overview}. When the step-size mismatch is observed, the renewal algorithm will be triggered to rectify step sizes automatically, ensuring a better training forward pass.

For the proposed theory and techniques, we conduct extensive experiments to validate the effectiveness. In comparison with prior arts, we demonstrate that our approaches can achieve superior accuracy while maintaining lower bit budgets, as shown in   Figure~\ref{fig:intro}.

In summary, our contributions are as follows:

\begin{itemize}
    \item We propose an adaptive bit allocation method to build high-performance SNNs with low bit budgets. To the best of our knowledge, this is the first work to realize an adaptive bit allocation for SNNs with direct training. 
    \item We accordingly refine the spiking neuron to exploit parametrization and temporal information, and achieve better experimental results during the adaptive bit allocation.
    \item We theoretically formulate the step-size mismatch issue that could severely harm the training of SNNs, and
    propose the step-size renewal mechanism to alleviate this issue, thus improving the overall model performance.
    \item We conduct thorough experimentation to demonstrate our proposed method's 
    effectiveness on both static and dynamic datasets. Comparative experiments also show that our model can achieve the advanced accuracy with lower bit budgets, as shown in  Figure~\ref{fig:intro}. 
\end{itemize}

\section{Related work}
\label{sec:related_work}

\subsection{Supervised direct learning of SNNs.}
Based on the idea that SNNs could be optimized end-to-end through back-propagation \citep{rumelhart1986learning}, Bothe  et~al. \citep{bohte2000spikeprop} first used the surrogate gradient (SG) to solve the non-differentiability of SNNs. Wu  et~al. \citep{wu2018spatio} further compared and analyzed the impacts of the shapes of SG functions on model performance. Afterward, many practical techniques and architectural designs were developed to strengthen the direct-trained SNN performance, such as tdBN \citep{zheng2021going}, SEW block \citep{fang2021deep}, spiking self-attention (SSA) block \citep{zhou2022spikformer}, TET \citep{deng2021temporal}, etc. In this paper, we adopt the two main-stream direct-trained SEW and SSA to estimate the effectiveness of our proposed methods.

\subsection{Multi-bit spiking neural networks.}
Wu  et~al. \citep{wu2021liaf} were the first to replace binary spikes with full-precision analog spikes to improve the model accuracy. Guo  et~al. added the spike amplitude coefficient \citep{guo2022real} and negative spike \citep{guo2024ternary} to realize ternary (2-bit) spikes. You  et~al. \citep{xing2024spikelm} and Xing  et~al. \citep{xing2024spikelm} concurrently introduced ternary spikes into natural language processing (NLP), and both achieved tremendous success. SpikeLLM \citep{xing2024spikellm} and Multi-bit SNN \citep{xiao2024multi} were more aggressive. They extended ternary spikes to multiple bits (over 2 bits) to boost the SNN performance while theoretically proving that the characteristic of addition-only computation can still be maintained. Shen  et~al. \citep{shen2024conventional} combined multi-bit SNNs with quantization and presented a rational and practical approach called "Bit Budget" to estimating computing overheads. Different from the above multi-bit SNNs, our work focuses on fine-grained bit allocations, with which multi-bit SNNs can substantially minimize the memory and computation and maximize the model accuracy.

\subsection{Mixed-precision quantization.}
Mixed-precision quantization is based on the insight that different layers in artificial neural networks (ANN) require different degrees of representation ability. Wang  et~al. \citep{wang2019haq} and Wu  et~al. \citep{wu2018mixed} were the pioneers that fostered the search-based means to allocate  different bit widths with reinforcement learning and differentiable neural architecture search, respectively. While, Dong  et~al. \citep{dong2019hawq} used Hessian metrics to represent the layer importance and accordingly assigned bit widths. Recently, optimization-based mixed-precision was developed to learn the bit width during training \citep{uhlich2019mixed,zhang2021differentiable,huang2022sdq,kim2024metamix}.
Compared with the above methods, which focus solely on ANN mixed-precision, our work is among the first to target SNNs, addressing the additional temporal dimension and mitigating step-size mismatch.

\section{Preliminaries}
\label{sec:preliminaries}

\subsection{Multi-bit spiking neuron}
Here, we begin by formulating the conventional Leaky Integrate-and-Fire (LIF) model in its discrete form:
\begin{align}
    &v^t_l=
    \begin{cases}\frac{1}{\tau}v^{t-1}_l+ V^t_{in,l}, S^{t-1}_{out,l}=0\\v_{rst},\text{otherwise}
    \end{cases}\label{eq:lif1}\\
&S^t_{out,l}=\begin{cases}1, v^t> V_{th} \\0, \text{otherwise}
\end{cases}\label{eq:lif2}
\end{align}
where $v^t_l$ denotes the membrane potential at time-step $t$ in the layer $l$ and can be updated with  Equation~\ref{eq:lif1}. $v_{rst}$ and $\tau$, respectively representing the resting potential and the time coefficient, are  constants. $V_{in,l}^t$, representing the input current, is a variable. In  Equation~\ref{eq:lif2}, $S^t_{out,l}$ is the binary spike whose value is determined by whether the membrane potential $v^t_l$ exceeds the membrane-potential threshold $V_{th}$. If that happened, $v^t_l$ would be reset to $v_{rst}$.
Plus, the input current $V_{in,l}^t$ is calculated from  the previous layer's $S^t_{out,l-1}$ via $V_{in,l}^t=\sum_jw_{l-1}S_{out,l-1}^t$, where $w_{l-1}$ denotes the weight parameters and $j$ denotes
the index of  spiking neurons.

Rooted from burst encoding \citep{liefficient}, the multi-bit spiking neuron generalizes  Equation~\ref{eq:lif2} to allow multiple spikes at the same time-step. The formulation transforms into \citep{shen2024conventional,xiao2024multi,xing2024spikellm}:
\begin{align}
     &V_{in,l}^t=\sum_j( w_{l-1}\cdot \alpha)S_{out,l-1}^t,\label{eq:mlif1}\\
&v^t_l=\frac{1}{\tau}v_l^{t-1}+ V^t_{in,l} - \alpha \cdot S^{t-1}_{out,l},\label{eq:mlif2}
\\
&S^t_{out,l}=\left \lfloor\frac{v^t_l}{V_{th}}\right \rfloor.\label{eq:mlif3}
\end{align}
Here, $\left \lfloor x \right \rfloor$ is the flooring function, and $\alpha$ is the spike amplitude coefficient.
With such transformation, the representation space of the spiking neuron is increased to $S^t_{out}\in \{0,1,..,2^{B_s}-1\}^T$. $B_s$ and $T$ denote the spike bit width and the temporal length, respectively.

\subsection{Quantization of weight parameters.} 
\label{sbsec:weight quantization}
For further improvement of SNN's efficiency, prior work has employed weight quantization to compress the model size, e.g., \citep{,shen2024conventional}. A classic quantization formula is given below:

\begin{align}
w^l_q= clip(\left \lfloor \frac{w_l}{S^l_q} \right \rceil, -2^{B_{w}-1}+1, 2^{B_{w}-1}-1 )\cdot S^l_q,\label{eq:weight quant}
\end{align}
where  $\lfloor x\rceil$ is the rounding function,  $S_q^l$ represents the quantization step size of layer $l$, $B_{w}$ denotes the weight bit width, $w_l$ is the weight parameter, and $w_q^l$ is the quantized weight value. If $B_{w}=1$, Equation~\ref{eq:weight quant} will switch to $w^l_q=sign(w_l/S^l_q)\cdot S^l_q$.

\subsection{Legitimate computation and memory measurement.}
\label{sbsec:sace nsace}
Shen et~al. \citep{shen2024conventional} propose the "Bit Budget" paradigm to fairly and practically estimate the substantial computation and memory. Following is the definition:
\begin{align}
    \text{Bit budget (BB)} = T \cdot B_{w} \cdot B_{s}.\label{eq:bit budget}
\end{align}
The corresponding  computational expenses evaluation, named Arithmetic Computation Effort (S-ACE), is defined by:
\begin{align}
    \text{S-ACE} = \sum_{w\in W, s\in S} n_{w,s}\cdot \text{BB} .\label{eq:sace}
\end{align}
$n_{w,s}$ is multiply-accumulate operations' number (MACs), $W$ and $S$ are weights and spikes' sets, respectively. 

{ 
To measure the computational consumption of fully ideal neuromorphic hardware, Shen et~al. \citep{shen2024conventional} also propose the NS-ACE metric to ignore the computation cost of non-spike data, which is calculated by:
\begin{equation}
\begin{split}
    \text{NS-ACE} &= \sum_{w\in W, s\in S} fr_s \cdot n_{w,s}\cdot \text{BB} \\
    &\approx \overline{fr}_s \cdot \sum_{w\in W, s\in S}   n_{w,s}\cdot \text{BB}\\
    &= \overline{fr}_s \cdot \text{S-ACE} .\label{eq:nsace}
\end{split}
\end{equation}
Here, $fr_s$ ($ \overline{fr}_s$ ) is the (average) firing rate of neurons, which is always 
less than $1$.
}

Notably, we regard the temporal length $T$ as a special "bit-width" since we have employed the concept of "Bit Budget". For conciseness, we will not explicitly distinguish between $T$ and 
other bit widths ($B_s, B_w$) in the subsequent context.

\section{Methodology}
\label{sec:methodology}
We build efficient and accurate SNNs by \textbf{(1)} quantizing SNNs, parametrizing bit widths, and refining the multi-bit spiking neuron formulation out of SNN's special necessity and better empiricism; \textbf{(2)} solving every non-differentiable term and designing  the loss function to achieve learnable and controllable bit allocation; \textbf{(3)} formulating the step-size mismatch issue and  alleviating it via the proposed step-size renewal.

\subsection{Parametrization}
\label{sbsec:parametrization}
% \subsection{Parametrization of bit widths} 
We can observe that  Equation~\ref{eq:mlif3} is essentially a flooring quantization process of $v^t_l$, where $S_{out,l}^t$ is the quantized integer, $V_{th}$ is the quantization step size, and the quantized spike value is $\alpha \cdot S_{out}^t$. Therefore, SNN's feature map size is controlled by $B_s$ and $T$. Similarly, the weight parameter size can also be controlled by the weight bit width $B_{w}$ in Equation~\ref{eq:weight quant}. Thus, the whole model's computation and memory overhead is controlled by $B_s$, $T$, and $B_w$, which is why the "Bit budget" metric \citep{shen2024conventional} is valid and matters. However, previous SNN quantization \citep{shen2024conventional} did not take into account a fine-grained allocation mechanism for these pivotal elements ($B_s$, $T$, and $B_w$), leading to suboptimal model efficiency.

Based on the insight that a layer with less representation ability but assigned with higher bit widths would incur memory and computation waste and interfere with the model inference \citep{wang2019haq,Chen_2021_ICCV}, we propose to parametrize the three pivotal elements layer-wise, making them learnable and can be optimized through back-propagation. Thus, a fine-grained allocation strategy would be feasible. Following are the parametrization formulas, including:
\begin{align}
    B_{s,l} = \left \lfloor clip(\hat{B}_{s,l},1,B_{s,bound})\right \rceil,\label{eq:spike bit param}
\end{align}
where $B_{s,bound}$ is the integer upper bound, which is manually set, $B_{s,l}$ is the spike bit width of layer $l$, and $\hat{B}_{s,l}$ is the learnable parameter of $B_{s,l}$. Similarly, 
the parametrization of $T$ and $B_w$ can be obtained:
\begin{align}
    &T_{l} = \left \lfloor clip(\hat{T}_{l},1,T_{bound})\right \rceil,\label{eq:spike len param}\\
    &B_{w,l} = \left \lfloor clip(\hat{B}_{w,l},1,B_{w,bound})\right \rceil.\label{eq:weight bit param}
\end{align}

As for the membrane-potential threshold $V_{th}$ and the weight quantization step size $S_q^l$, no specific parametrization formulas are introduced since they inherently exist as floating-point values. 

Consequently, different layers can attain suitable bit widths and temporal lengths for different data, leading to adaptive and flexible allocation of memory and computation.  

\subsection{Formulation of the refined parametric spiking neuron} 
With the quantization and parametrization implemented, the original multi-bit spiking neuron is unsuitable for the following bit-width learning process. Therefore, we refine its formulation out of necessity and empiricism.

Firstly, since the temporal length is made parametric and can be different among different layers during
training, two issues arise: \textbf{(1)} the gradient requirement, i.e., \emph{$T_{l}$ needs to get involved in the forward pass}; \textbf{(2)} the temporal mismatch between layers. 
For example, the previous layer yields a spike train with $T=2$, while the post layer has four temporal bits, i.e., $T=4$. 
Two time-steps of data remain unfilled. 
To solve these issues, we exploit the temporal flexibility \citep{yao2021temporal}, directly squeezing the spike train through an averaging operation, as shown in Figure~\ref{fig:overview}(a) and  Equation~\ref{eq:our sn1}. Notably, the extra computation brought by such temporal squeezing is negligible because it only makes up less than 0.1\% of the total model computation. In addition, the spiking self-attention\citep{zhou2022spikformer} layer requires the three  spike feature maps ($Q$, $K$, and $V$) to perform matrix multiplication, necessitating compatible feature map dimensions. Therefore, we enforce shared $T_l$ among the corresponding spiking neurons to ensure the intra-layer temporal matching, as shown in Figure~\ref{fig:overview}(a).

Secondly, the flooring operation in  Equation~\ref{eq:mlif3} may induce extra quantization error that incurs model degradation. 
From a value quantization perspective, compared to the rounding operation, the flooring operation brings in more quantization errors, which is evident.  
The proof is given in the appendix. 
Therefore, we offer another choice: a membrane-potential shift can be added to the flooring operation of Equation~\ref{eq:mlif3}, making Equation~\ref{eq:mlif3} equivalent to the rounding operation when the shift value is proper. Meanwhile, the original spike-fire format has also been maintained. The transformed Equation~\ref{eq:mlif3} is Equation~\ref{eq:our sn3}. 
For the same reason, we find the time constant $\tau$, apart from increasing stochasticity, is implausible. Therefore, setting $\tau=1$ should also be considered.
Based on the above insights, we conduct experiments in  Table~\ref{exp:ablation on refined neuron} as empirical support.

Thirdly, the spike amplitude coefficient $\alpha$ is usually defined differently among prior arts  \citep{shen2024conventional, guo2024ternary, xiao2024multi}. Here, we keep consistent with quantization's perspective, and directly unify $V_{th}$ and $\alpha$ as identical, layer-wise, learnable quantization step size, as shown in Equations \ref{eq:our sn1}- \ref{eq:our sn3}.

Eventually, we formulate our refined multi-bit spiking neuron in the following and illustrate it in  Figure~\ref{fig:overview}.
\begin{align}
    &V_{in,l}^t=\sum_j w^{l-1}_q \cdot V^1_{th,l-1}\cdot \frac{1}{T_{l-1}} \sum_t^{T_{l-1}}  S_{out,l-1}^t,\label{eq:our sn1}\\
&v^t_l=\frac{1}{\tau}v_l^{t-1}+ V^t_{in,l}- S^{t-1}_{out,l}V^1_{th,l},\label{eq:our sn2}
\\
&S^t_{out,l}= clip(\left \lfloor\frac{v^t_l}{V^{1,t}_{th, l}}+V^{2,t}_{th, l}\right \rfloor,  0, 2^{B^t_{s,l}}-1 ),\label{eq:our sn3}
\end{align}
where, in our practice, $\tau$ is suggested to be 1, the spike amplitude coefficient $\alpha$ is replaced by the learnable membrane-potential threshold $V^1_{th,l}$, and  threshold shift $V^2_{th, l}$ equates to  $0.5sign(v^t_l/V^1_{th,l})$. 
Beyond this, to further increase the adaptivity \citep{yao2022glif, xing2024spikelm}, we apply the temporal-wise sharing technique to $V_{th,l}^1$ and $B_{s,l}$.
Following \citep{yao2022glif} and \citep{xing2024spikelm}, we make $V_{th,l}^1$ and $B_{s,l}$ shared temporally along the temporal dimension, as shown in  Equation~\ref{eq:our sn3}. 
Specifically, 
$
        V_{th,l}^1 \Rightarrow V_{th,l}^{1,t} 
        \in \{V_{th,l}^{1,1},..., V_{th,l}^{1,t},..., V_{th,l}^{1,T_{bound}}\}. 
$
Similarly, 
$
        B_{s,l} \Rightarrow B_{s,l}^{t} 
        \in \{B_{s,l}^{1},..., B_{s,l}^{t},..., B_{s,l}^{T_{bound}}\}. 
$

As suggested in \citep{guo2022real,guo2024ternary, liu2022spikeconverter}, negative spikes should be enabled to improve the representation ability of SNNs. Therefore, we expand Equation~\ref{eq:our sn3} to the bidirectional domain, thus supporting negative spikes. Equation~\ref{eq:our sn3} would turn into:
\begin{equation}
S^t_{out,l}= clip(\left \lfloor\frac{v^t_l}{V^{1,t}_{th, l}}+V^{2,t}_{th, l}\right \rfloor,  -2^{B^t_{s,l}-1}+1, 2^{B^t_{s,l}-1}-1 ).\label{eq:our bsn3}
\end{equation} 
Notably, in our practice, we find that for both spiking ResNet and Spikformer, only the very first convolution layer would require such bidirectional spiking neuron to encode the RGB image signals or the bipolar event signals. At the same time, other layers keep using the positive multi-bit spiking neuron. As a result, models would achieve higher accuracy.

\subsection{Adaptive bit width} 
\label{sbsec:bit learn}
With parametrized bit widths $B_{s,l}$, $B_{w,l}$ and $T_l$, and learnable quantization step sizes $V^1_{th,l},S^l_q$ as illustrated in the above subsections, this subsection further solves the non-differentiable terms appearing in the parametrization process and  introduces the regulating loss, which regulates bit widths to converge to the targets during training.

\subsubsection{Gradient calculations} 
\label{sbsec:grad cal}
We solve non-differentiation via straight-through estimator (STE) \citep{bengio2013estimating} and chain rule. 
% The derivation process is deferred to the appendix. 
\\\textbf{1)} For  $B_{s,l}$, 
\begin{align}
    &\frac{\partial L_{task}}{\partial \hat{B}_{s,l}} = \frac{\partial L_{task}}{\partial B_{s,l}}= \sum^T_t \sum_j \frac{\partial L_{task}}{\partial S^t_{out,l}}  \cdot g_{s,scale}\cdot g_{q}, \\ 
&g_{q} =
\begin{cases}
sign(\frac{v^t_l}{V_{th,l}^1})   \cdot (q_{max}+1)   \cdot    \ln2   , \frac{v^t_l}{V_{th,l}^1}>q_{max} \\0, \text{otherwise}.
\end{cases}
\end{align}
Here, $j$ is the index of the feature map element. The minimum quantization integer $q_{min}=0$ and the maximum integer $q_{max}=2^{B_{s,l}^t}-1$. While, $g_{s,scale}$, equal to $\frac{1}{\sqrt{\sum_j1\cdot q_{max}}}$ and meant to avoid gradient explosion \citep{esser2020learned}, is the gradient scaling factor. $L_{task}$ represents the task loss signal.
\\\textbf{2)} For  $B_{w,l}$,
\begin{align}
    &\frac{\partial L_{task}}{\partial \hat{B}_{w,l}} = \frac{\partial L_{task}}{\partial B_{w,l}}=  \sum_i \frac{\partial L_{task}}{\partial w^l_q}  \cdot g_{w,scale} \cdot g_{q}, \\ 
&g_{q} =
\begin{cases}
sign( \frac{w_l}{S^l_q})  S^l_q \cdot (q_{max}+1)   \cdot    \ln2   , \frac{w_l}{S^l_q}\notin[q_{min}, q_{max}] \\0, \text{otherwise}.
\end{cases}
\end{align}
Specially, when $B_{w,l}=1$, $q_{min}=-1$ and $q_{max}=1$. Otherwise, $q_{min}=-2^{B_{w,l}-1}+1$ and $q_{max}=2^{B_{w,l}-1}-1$. Similarly,  $i$ is the index of the weight element, and $g_{w,scale} = \frac{1}{\sqrt{\sum_i1\cdot q_{max}}}$.
% \\\textbf{3)} For  $V^1_{th,l}$,
% \begin{align}
% &\frac{\partial L_{task}}{\partial V^1_{th,l}} =  \sum^T_t \sum_j \frac{1}{V^1_{th,l}} \frac{\partial L_{task}}{\partial S^t_{out,l}}  \cdot g_{s,scale}  \cdot g_{q}, \\ 
% &g_{q} =
% \begin{cases}
% \left\lfloor\frac{v^t_l}{V^1_{th,l}}+V_{th,l}^2\right\rfloor - \frac{v^t_l}{V^1_{th,l}}, \frac{v^t_l}{V^1_{th,l}} \in [q_{min},q_{max}]\\
% 0, \frac{v^t_l}{V^1_{th,l}}<q_{min}\\
% q_{max}, q_{max}<\frac{v^t_l}{V^1_{th,l}},
% \end{cases}
% \end{align}
% where all the notations are the same to the case \textbf{1)}.
% \\\textbf{4)} For  $S^l_q$,
% \begin{align}
% &\frac{\partial L_{task}}{\partial S_q^l} =  \sum_i \frac{\partial L_{task}}{\partial w^l_q} \cdot g_{w,scale} \cdot g_{q}, \\ 
% & g_{q}= 
% \begin{cases}
% \left \lfloor \frac{w_l}{S^l_q} \right\rceil - \frac{w_l}{S^l_q}, \frac{w_l}{S^l_q} \in [q_{min}, q_{max}]\text{ and }B_{w,l}>1\\
% sign(\frac{w_l}{S^l_q}) - \frac{w_l}{S^l_q}, \frac{w_l}{S^l_q}\in [q_{min}, q_{max}]\text{ and }B_{w,l}=1\\
% q_{min}, \frac{w_l}{S^l_q}<q_{min}\\
% q_{max}, q_{max}<\frac{w_l}{S^l_q},
% \end{cases}
% \end{align}
% where, notations are all the same to the case \textbf{2)}.
\\\textbf{3)} For $V^1_{th,l}$ and $S^l_q$, the derivation of gradients is similar to prior arts \citep{esser2020learned} and trivial, so we defer it to the appendix.
\\\textbf{4)} For $S^t_{out,l}$ and $T_l$, the only non-differentiable terms are the flooring and clipping of  Equation~\ref{eq:our sn3}. We solve them by:
\begin{align}
    \frac{\partial S^t_{out,l}}{\partial (v^t_l/V_{th,l}^1 )} = 
\begin{cases}
1, v^t_l/V_{th,l}^1 \in [q_{min}, q_{max}]\\
0,\text{otherwise}.
\end{cases}
\end{align}
$q_{min}$ and $q_{max}$ are in line with the case \textbf{1)}. Thus, $\frac{\partial L_{task}}{\partial S^t_{out,l}}$ and $\frac{\partial L_{task}}{\partial \hat{T}_l}$ can be easily derived via the trivial chain rule.

\subsubsection{Loss function} 
\label{sbsec:loss function}
With the above gradient calculations, 
the bit-width parameters $\hat{B}_{s,l}$, $\hat{B}_{w,l}$, and $\hat{T}_l$ can directly obtain the task loss signal $L_{task}$ for optimization. However,  $L_{task}$ alone does not guarantee that these bit-width parameters can be compressed towards the target bit widths. 
Therefore, we exploit a direct constraint on the average bit width of the model in the following:
\begin{equation}
\begin{split}
& L_{total}=L_{task}+\lambda_1L(\bar{B}_w, B^{tar}_w)+\lambda_2L(\bar{T}, T^{tar})\\
&+\lambda_3L(\bar{B}_s, B^{tar}_s), \text{ where } L(x,y)=|| x-y||_2^2.
\end{split}\label{eq:loss function}
\end{equation}
\( B^{tar}_w \), \( T^{tar} \), and \( B^{tar}_s \) represent the target bit widths. \( \lambda_1 \), \( \lambda_2 \), and \( \lambda_3 \) are penalty coefficients. \( \bar{B}_w \) is the average bit width of each weight parameter. \( \bar{T} \) and \( \bar{B}_s \) are feature element's average time-step and bit width, respectively.

\begin{figure*}[t]
  \centering
  \includegraphics[width= 0.9\textwidth]{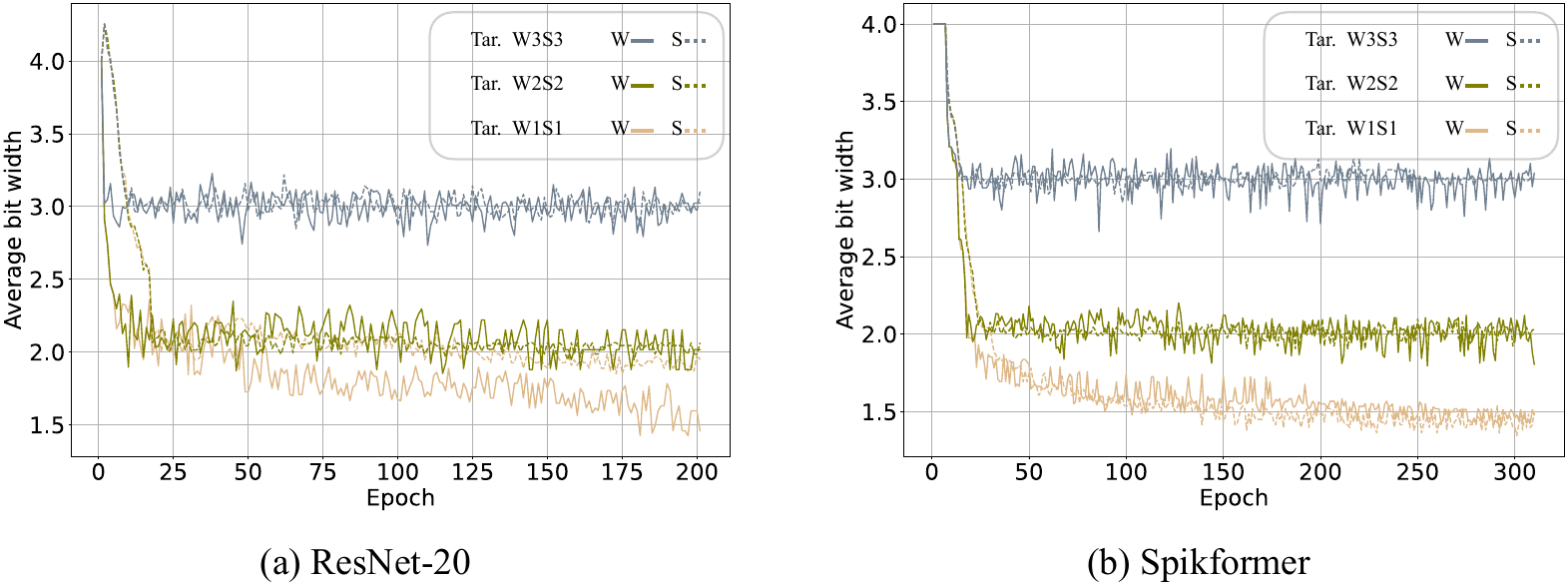}
  \caption{Average bit width changes of ResNet-20 and Spikformer on CIFAR-10 during the adaptive-bit-width training. Tar. abbreviates target bit width. W and S denote weight and spike bit width, respectively.
  }
  \label{fig:bit detect}
\end{figure*}

\subsection{Quantization step-size renewal}
\label{sbsec:renew}
\subsubsection{Quantization step size mismatch issue} 
Bit width is changeable due to parametrization. If it decreases after the previous parameter update, the current quantization step size in the next forward pass will become mismatched. Because the change of bit width belongs to integer mutation while step size does not. Consequently, the quantization error would be increased. 

Assuming a variable $x$, e.g., the activation after convolution, submits to $N(0,\sigma^2)$ and is then filtered via ReLU. Then, only $x>0$ needs to be considered because $x=0$ will not incur any quantization error.  We get $x\sim 2N(0,\sigma^2)| x>0$. Thus, we can offer a theoretical formulation and analysis of the step-size mismatch issue.

\begin{theorem} 
In the quantization process of the variable $x$: $x_q = s\cdot clip(\left\lfloor\frac{x}{s}\right\rceil,0,2^b-1)$, where $x\sim 2N(0,\sigma^2)|  x>0$; $b$ and $s$ respectively denote the bit width and the quantization step size.  
Let the quantization step size $s$ be the statistically optimal: $3\sigma=s\cdot(2^b-1)\Rightarrow s=\frac{3\sigma}{2^b-1}$ and the quantization error $Err=(x-x_q)^2$. If $b$ decreases to $b'$, the quantization error would  
increase by the possibility of $P(x>\frac{2^{b'}-1}{2^{b}-1}\cdot3\sigma| (x\sim 2N(0,\sigma^2),  x>0)) $. 
(Theorem and proof of the bidirectional domain, i.e., no ReLU applied and $x\sim N(0,\sigma^2)$, are in the appendix.)
\end{theorem}

\begin{pf} 
Given $b'<b$ and $x'_q=s\cdot  clip(\left\lfloor\frac{x}{s}\right\rceil,0,2^{b'}-1)$, we obtain $x'_q = x_q - \lambda$, where  $\lambda=s[clip(\left\lfloor\frac{x}{s}\right\rceil,2^{b'}-1,2^b-1)-(2^{b'}-1)]$. Thus, $Err'=(x-x_q+\lambda)^2$, and the possibility of $P(Err'>Err)$ equates to $P(\lambda > 0)$ (the deduction here is deferred to the appendix). We can get 
\begin{equation}
\begin{split}
P(\lambda>0) &= P(\frac{x}{s}>2^{b'}-1)\\
&=P(x>\frac{2^{b'}-1}{2^{b}-1}\cdot 3\sigma).
\end{split}
\end{equation}
\end{pf}

We conclude that \emph{\textbf{I)} a larger decrease in bit width leads to a higher possibility of increased quantization error.}

\begin{lemma}
Because $b'\le b-1$ and $b'\ge1$, we get $\frac{2^{b'}-1}{2^{b}-1}\le \frac{1}{2}$. Thus, the lower bound of $P(\lambda>0)$ can be calculated:
\begin{equation}
\begin{split}
    P(\lambda>0) > P(x>\frac{3\sigma}{2} ).
\end{split}
\end{equation}
\end{lemma}

Since $x\sim 2N(0,\sigma^2)$ and $x>0$, we obtain $P(x>\frac{3\sigma}{2} ) = 0.1336>0$. Consider a tensor $X=\{x_1,x_2,...\}^n$, we can conclude that \emph{\textbf{II)}
as bit width decreases, the unchanged step
size would inevitably incur more quantization errors when $X$ has a large dimension $n$.}

With the conclusion \emph{\textbf{I)}} and \emph{\textbf{II)}} combined, 
the step-size mismatch issue can be proven  harmful in direct training, where bit-width mutation constantly occurs between forward passes as shown in  Figure~\ref{fig:bit detect}. As the model goes deeper, quantization error accumulates and seriously impairs the model output \citep{sengupta2019going}. \emph{The step size mismatch influence could be more severe in SNNs because the invalid step size will be consecutively used along the temporal dimension in  Equation~\ref{eq:mlif3}.} Therefore, the quantization error will accumulate temporally, causing serious inference offsets. For instance, if $X = \{x_1, x_2,...,x_T\}$, then the possibility of a increased quantization error $(X-X_q)^2$ would be $1-[1-P(x>\frac{2^{b'}-1}{2^b-1}3\sigma)]^T$, which increases as $T$ rises.

\subsubsection{Step-size renewal mechanism} 

\begin{algorithm}[t]
	% \textsl{}\setstretch{1.8}
	\renewcommand{\algorithmicrequire}{\textbf{Input:}}
	\renewcommand{\algorithmicensure}{\textbf{Output:}}
    \caption{Step-size renewal algorithm.}
    \label{algo:step-size renewal}
    \textbf{Initialize the bit-width observer:} The running maximum $V_{r\_{max}}$ is set to $-\infty$ and the minimum $V_{r\_{min}}$ is set to $\infty$. The recorded bit width $B_l'$ is set to $0$. 
    \begin{algorithmic}[1]
		\REQUIRE Data to be quantized $X$ and $X$'s current bit width $B_l$.

		\ENSURE  The optimal $S_l$.
        % \STATE   \# The bit-width observer.
        \STATE   if $B_{l}\ne B'_{l}$ do
        \STATE  \quad $B'_{l}\gets B_{l}$ 
        \STATE  \quad if $X$ is defined in the bidirectional domain do
        \STATE  \quad \quad $q_{min}\gets-2^{B'_{l}-1}+1$  
        \STATE  \quad \quad $q_{max}\gets2^{B'_{l}-1}-1$
        \STATE  \quad else do
        \STATE  \quad \quad $q_{min}\gets 0$  
        \STATE  \quad \quad $q_{max}\gets 2^{B'_{l}}-1$
        \STATE  \quad $V_{max},V_{min} \gets max(X), min(X)$ 
        \STATE  \quad $V'_{max}$, $V'_{min}$ $\gets$ Input $(X,V_{min},V_{max})$ to Algorithm~\ref{algo:grid search}.
        \STATE  \quad $V_{r\_max},V_{r\_min}, \gets max(V'_{max},V_{r\_max}),min(V'_{min},V_{r\_min})$ 
        \STATE  \quad $S_l \gets \frac{V_{r\_max}-V_{r\_min}}{q_{max}-q_{min}}$
        \STATE  \quad return $S_l$
        \STATE else do
        \STATE  \quad   pass
    \end{algorithmic}
\end{algorithm}

\begin{algorithm}[t]
	% \textsl{}\setstretch{1.8}
	\renewcommand{\algorithmicrequire}{\textbf{Input:}}
	\renewcommand{\algorithmicensure}{\textbf{Output:}}
    \caption{Grid search on $V_{max}$ and $V_{min}$.}
    \label{algo:grid search}
    \begin{algorithmic}[1]
		\REQUIRE Data to be quantized $X$, the $V_{max}$ and $V_{min}$ of $X$, the iteration number $K$, the minimum quantization integer $q_{min}$, the maximum quantization integer $q_{max}$, and the power constant $pow$.
  
		\ENSURE The optimal $V'_{max}$ and $V'_{min}$.

        \STATE $Score\gets 0$, $R\gets V_{max}-V_{min}$, $V'_{max} \gets V_{max}$, and $V'_{min} \gets V_{min}$.

        \STATE for $k=1; k\le K; i++$ do

        \STATE    \quad $V_{max}\gets \frac{k\cdot R}{K} $
        \STATE    \quad if $any(X<0)$ do
        \STATE    \quad \quad $V_{min}\gets -\frac{k\cdot R}{K} $
        \STATE    \quad else do
        \STATE    \quad \quad $V_{min}\gets 0 $
        \STATE  \quad $S'_l\gets\frac{V_{max}-V_{min}}{q_{max}-q_{min}}$ 

        \STATE  \quad $X_{q,k}\gets S'_l\cdot  clip(\left\lfloor\frac{X}{S'_l}\right\rceil,q_{min},q_{max})$
        \STATE  \quad $Score' \gets mean(|X_{q,k}-X|^{pow}) $
        \STATE  \quad if $Score' < Score$ do
        \STATE  \quad \quad $Score\gets Score'$
        \STATE  \quad \quad $V'_{max}, V'_{min} \gets V_{max}, V_{min}$ 
        \STATE return  $V'_{max}$, $V'_{min}$
    \end{algorithmic}$mean(x)$ is the element-wise averaging function. $any(x)$ checks if at least one element is True.
\end{algorithm}

To alleviate the mismatch issue, we propose the following step-size renewal. 
As illustrated in  Figure~\ref{fig:overview},
the renewal mechanism is based on a bit-width observer that watches on the bit width $B_l$, e.g., $B_{w,l}$ and $B_{s,l}$, and records the running maximum $V_{r\_max}$ and minimum $V_{r\_min}$ of the  data to be quantized. $V_{r\_max}$ and $V_{r\_min}$, reflecting the overall data distribution, will be used to renew the step size $S_l$, e.g., $S_q^l$ and $V_{th,l}^1$, directly.

As illustrated in Algorithm~\ref{algo:step-size renewal}, once $B_l$ is changed, the new $B_l$ will be recorded, the new $q_{max}$ and $q_{min}$ are calculated accordingly, and the observer will also instantly read the current maximum $V_{max}$ and minimum $V_{min}$ of the current batch of data $X$, e.g., weight or activation. 
Based on $V_{max}$ and $V_{min}$, a grid search is performed to calculate the currently optimal maximum $V'_{max}$ and minimum $V'_{min}$  as written in Algorithm~\ref{algo:grid search}.  Finally, the running maximum $V_{r\_max}$ and minimum $V_{r\_min}$ are updated via 
\begin{equation}
\begin{split}
    &V_{r\_max} = max(V'_{max},V_{r\_max}),\\&V_{r\_min} = min(V'_{min},V_{r\_min}). 
\end{split}
\end{equation}
And, the renewed step size $S_l$ is determined via
\begin{align}
    S_l=\frac{V_{r\_max}-V_{r\_min}}{q_{max}-q_{min}}.
\end{align}

Thus, an optimal $S_l$ can be applied and cover the original one to correct the current forward pass. 
Since the grid search process will consume extra training time, we only activate the renewal mechanism in the drastic bit reduction stage. Based on the observations of  Figure~\ref{fig:bit detect}, bit reduction mainly occurs in the first 12.5\% epochs. Therefore, we set a shutting threshold that once the difference between the current and the target bit width is below 24\% of the initial difference, the renewal mechanism would be deactivated. Consequently, the training time increased by the grid search is lower than 0.05\% of the total training time.

\begin{table*}[width=1.\textwidth,cols=10,pos=t]
\centering
\caption{Comparisons with existing works on CIFAR. W/S/T denotes the averaged weight/spike/temporal bit width. U-Quant., following  \citep{shen2024conventional}, is the plain uniform quantization on SNN, which uses fixed uniform bit widths. Our models are initial to W/S/T=4/4/2.}
\label{tab:comparison on cifar}
\begin{adjustbox}{max width=1.\textwidth}
\begin{threeparttable}
\begin{tabular}{cccccccccccc}
\toprule
\multirow{2}{*}{\textbf{Method}}                                        & \multirow{2}{*}{\textbf{Architecture}} & \multicolumn{5}{c}{\textbf{CIFAR 10}}                                                                         & \multicolumn{5}{c}{\textbf{CIFAR 100}}                                                                   \\ \cmidrule{3-12} 
                                                                        &                                        & \textbf{W/S/T}          & \textbf{Bit Budget} & \textbf{S-ACE (G)}      & \textbf{Size (MB)} & \textbf{Top-1} & \textbf{W/S/T}          & \textbf{Bit Budget} & \textbf{S-ACE (G)}      & \textbf{Size (MB)}     & \textbf{Top-1} \\ \midrule
\multirow{2}{*}{Dspike \citep{li2021differentiable}}   & \multirow{2}{*}{ResNet18}              & 16/1/2                  & 32                  & 17.72                   & 22.42              & 93.13          & 16/1/2                  & 32                  & 17.72                   & 22.42         & 71.68          \\
                                                                        &                                        & 16/1/4                  & 64                  & 35.44                   & 22.42              & 93.66          & 16/1/4                  & 64                  & 35.44                   & 22.42         & 73.35          \\
MLF \citep{fengmulti}                                  & ResNet20                               & 16/1/4                  & 64                  & 165.58                  & 34.55              & 94.25          & 16/1/4                  & 64                  & 165.58                  & 34.55         & -              \\
\multirow{2}{*}{tdBN \citep{zheng2021going}}           & \multirow{2}{*}{ResNet19}              & 16/1/2                  & 32                  & 73.28                   & 25.25              & 92.34          & 16/1/2                  & 32                  & 73.28                   & 25.25         & -              \\
                                                                        &                                        & 16/1/4                  & 64                  & 146.56                  & 25.25              & 92.92          & 16/1/4                  & 64                  & 146.56                  & 25.25         & -              \\
\multirow{2}{*}{TET \citep{deng2021temporal}}          & \multirow{2}{*}{ResNet19}              & 16/1/2                  & 32                  & 73.28                   & 25.25              & 94.16          & 16/1/2                  & 32                  & 73.28                   & 25.25         & 72.87          \\
                                                                        &                                        & 16/1/4                  & 64                  & 165.12                  & 25.25              & 94.44          & 16/1/4                  & 64                  & 165.12                  & 25.25         & 74.47          \\ \midrule
\multirow{3}{*}{Ternary Spike \citep{guo2024ternary}}  & ResNet19                               & 16/2/1                  & 32                  & 330.24                  & 25.25              & 95.58          & 16/2/1                  & 32                  & 330.24                  & 25.25         & 78.45          \\ \cmidrule{2-12} 
                                                                        & \multirow{2}{*}{ResNet20}              & 16/2/2                  & 64                  & 165.58                  & 34.55              & 94.48          & 16/2/2                  & 64                  & 165.58                  & 34.55         & 73.41          \\
                                                                        &                                        & 16/2/4                  & 128                 & 330.24                  & 34.55              & 94.96          & 16/2/4                  & 128                 & 330.24                  & 34.55         & 74.02          \\ \midrule
\multirow{3}{*}{Multi-bit Spike \citep{xiao2024multi}} & \multirow{3}{*}{ResNet20}              & 16/4/1                  & 64                  & 165.58                  & 34.55              & 94.59          & 16/4/1                  & 64                  & 165.58                  & 34.55         & 75.43          \\
                                                                        &                                        & 16/4/4                  & 256                 & 660.48                  & 34.55              & 94.93          & 16/4/4                  & 256                 & 660.48                  & 34.55         & 76.51          \\
                                                                        &                                        & 16/4/6                  & 384                 & 990.72                  & 34.55              & 95.00          & 16/4/6                  & 384                 & 990.72                  & 34.55         & -              \\ \midrule
\multirow{2}{*}{U-Quant.$\dagger$}                                      & \multirow{2}{*}{ResNet20}              & 4/4/1                   & 16                  & 41.40                   & 8.64               & 96.35          & 4/4/1                   & 16                  & 41.40                   & 8.64          & 79.83          \\
                                                                        &                                        & 4/4/2                   & 32                  & 82.79                   & 8.64               & 96.39          & 4/4/2                   & 32                  & 82.79                   & 8.64          & 80.19          \\ \midrule
Ours-111                                                                & \multirow{2}{*}{ResNet18}              & \textbf{1.45/1.61/1.04} & \textbf{2.43}       & \textbf{2.96}           & \textbf{2.03}      & \textbf{95.13} & \textbf{1.65/1.87/1.03} & \textbf{3.17}       & \textbf{4.29}           & \textbf{2.32} & \textbf{75.42} \\
Ours-222                                                                &                                        & \textbf{1.97/2.04/2.01} & \textbf{8.1}        & \textbf{11.59 / 6.56*}   & \textbf{2.75}      & \textbf{96.01} & \textbf{2.06/2.08/2.02} & \textbf{8.69}       & \textbf{11.34 / 6.86*}  & \textbf{2.89} & \textbf{75.80} \\ \midrule
Ours-111                                                                & \multirow{3}{*}{ResNet20}              & \textbf{1.38/1.32/1.00} & \textbf{1.83}       & \textbf{7.55}           & \textbf{2.98}      & \textbf{96.15} & \textbf{2.06/1.58/1.00} & \textbf{3.25}       & \textbf{13.37}          & \textbf{4.45} & \textbf{79.70} \\
Ours-222                                                                &                                        & \textbf{2.13/2.00/2.08} & \textbf{8.87}       & \textbf{29.75 / 15.93*} & \textbf{4.60}      & \textbf{96.48} & \textbf{2.11/1.98/2.08} & \textbf{8.66}       & \textbf{32.67 / 18.83*} & \textbf{4.56} & \textbf{80.59} \\
Ours-341                                                                &                                        & \textbf{3.01/4.00/1.00} & \textbf{12.08}      & \textbf{39.49}          & \textbf{6.51}      & \textbf{97.02} & \textbf{2.88/4.04/1.00} & \textbf{11.64}      & \textbf{40.60}          & \textbf{6.23} & \textbf{81.30} \\ \bottomrule
\end{tabular}
\begin{tablenotes}
\footnotesize
\item $\dagger$ denotes plain uniform quantization, following  \citep{shen2024conventional}. $*$ denotes the temporal squeezing is applied in inference.  "Ours-xxx" denotes the target bit widths of W/S/T. 
\end{tablenotes}
\end{threeparttable}
\end{adjustbox}
\end{table*}

\begin{table*}[width=1.\textwidth,cols=7, pos=t]
\centering
\caption{Comparisons with existing works on ImageNet.}
\label{tab:comparison on imagenet}
\begin{adjustbox}{max width=\textwidth}
\begin{threeparttable}\begin{tabular}{cccccccc}
\toprule
\textbf{Method}                                                 & \textbf{Architecture}         & \textbf{W/S/T}         & \textbf{Bit Budget} & \textbf{S-ACE (G)} & {\textbf{NS-ACE(G)}} & \textbf{Size (MB)} & \textbf{Top-1} \\ \midrule
Hybrid training \citep{rathi2020enabling}      & ResNet34                      & 16/1/250               & 4000                & 13608              & -                  & 43.56              & 61.48          \\
Spiking ResNet \citep{hu2021spiking}           & ResNet50                      & 16/1/350               & 5600                & 18952              & -                  & 51.12              & 72.75          \\
tdBN \citep{zheng2021going}                    & ResNet34                      & 16/1/6                 & 96                  & 340.39             & -                  & 43.56              & 63.72          \\
Dspike \citep{li2021differentiable}            & ResNet34                      & 16/1/6                 & 96                  & 340.39             & -                  & 43.56              & 68.19          \\
MS-ResNet \citep{hu2024advancing}              & ResNet34                      & 16/1/6                 & 96                  & 340.39             & -                  & 43.56              & 69.42          \\
Attention-SNN \citep{yao2023attention}         & ResNet34                      & 16/1/1                 & 16                  & 56.73              & -                  & 43.56              & 69.15          \\ \midrule
TET \citep{deng2021temporal}                   & Sew-ResNet34                  & 16/1/4                 & 64                  & 226.93             & -                  & 43.56              & 68.00          \\
Ternary Spike \citep{guo2024ternary}           & ResNet34                      & 16/2/4                 & 128                 & 435.86             & -                  & 43.56              & 70.74          \\
\multirow{2}{*}{PLIF+SEW \citep{fang2021deep}} & SEW-ResNet34                  & 16/1/4                 & 64                  & 226.93             & {63.14}             & 43.56              & 67.04          \\
                                                                & SEW-ResNet50                  & 16/1/4                 & 64                  & 215.62         & {69.03}              & 51.12              & 67.78          \\
Quant. SNN  \citep{shen2024conventional}        & SEW-ResNet34                  & 8/8/1                  & 64                  & 226.93             & {56.42}             & 21.78              & 70.13          \\ \midrule
Ours-221                                                        & \multirow{3}{*}{SEW-ResNet34} & \textbf{2.58/2.18/1.0} & \textbf{5.61}       & \textbf{30.74}     & { \textbf{7.77}}      & \textbf{7.03}      & 70.57          \\
Ours-241                                                        &                               & \textbf{2.62/3.96/1.0} & \textbf{10.39}      & \textbf{54.69}     & { \textbf{16.99}}    & \textbf{7.14}      & \textbf{72.02} \\
Ours-441                                                        &                               & \textbf{3.93/3.92/1.0} & \textbf{15.4}       & 77.78              & { \textbf{24.30}}    & \textbf{10.69}     & \textbf{72.82} \\ \bottomrule
\end{tabular}

% \begin{tablenotes}
% \footnotesize
% \item $\dagger$ denotes plain uniform quantization, following  \citep{shen2024conventional}.
% \item $*$ denotes the temporal squeezing is applied in inference.
% \end{tablenotes}
\end{threeparttable}
\end{adjustbox}
\end{table*}

\section{Experiments}
\label{sec:experiments}
In this section, we first compare our methods with the existing works using ResNet  on CIFAR \citep{krizhevsky2009learning} and ImageNet-1k \citep{deng2009imagenet}, then
conduct ablation studies to verify different aspects of our methods, including \textbf{(1)} each proposed technology's verification, \textbf{(2)} our methods' extension to spiking self-attention \citep{zhou2022spikformer}, \textbf{(3)} mitigating the "temporal constraint" of the prior arts \citep{shen2024conventional} on the event dataset CIFAR10-DVS \citep{li2017cifar10} and DVS-GESTURE \citep{amir2017low}, 
\textbf{(4)} the learned bit allocation's visualization, and \textbf{(5)} comparing the mixed-precision ANNs. All implementation details are attached in the appendix.

\subsection{Comparisons with existing works}
\label{sbsec:comparison}

\subsubsection{CIFAR} We apply our methods to  ResNet20 \citep{guo2024ternary} and ResNet18 \citep{li2021differentiable} to compete with the advanced SNNs. As shown in  Table~\ref{tab:comparison on cifar}, we first compare our methods with those using extra learning- or module-based model optimization \citep{li2021differentiable, deng2021temporal, zheng2021going}, and achieve superior accuracy using much lower memory (bit budget) and computation (S-ACE). In the case of ResNet-18 with target W/S/T being 1/1/1, our methods can also compress the model sizes to one-tenth of their original dimensions, while achieving at most 2\% and 3.74\% higher model accuracy on CIFAR-10 and CIFAR-100, respectively. Such a reduction of storage requirements would make SNNs more compatible with on-chip computing architectures \citep{akopyan2015truenorth, mao2024stellar}.
Comparing with Ternary Spike ResNet20 \citep{guo2024ternary} that adopts 2-bit bidirectional spike, we can outperform it by at least 1.19\% and 5.88\% accuracy increase on CIFAR-10 and -100, respectively, while using average 1.32- and 1.58-bit spike and reducing 91.37\% and 90.59\% model sizes. 
Then, we compare with Multi-bit ResNet20 \citep{xiao2024multi}, whose memory and computation respectively increase to huge 256 bit budgets and 660.48 G S-ACEs. While, our model can achieve not only 1.22\% and 3.39\% accuracy increase on CIFAR-10 and CIFAR-100 but also two magnitudes of memory and computation savings.
Finally, compared with plain uniform quantization (U-quant.), our methods can maximize the model accuracy while minimizing memory and computation requirements. The advantages of our approaches in terms of memory usage, computational efficiency, and storage requirements are consistently demonstrated.
Additionally, since we adopt temporal squeezing as shown in  Figure~\ref{fig:overview} and Equation~\ref{eq:our sn1}, the temporal dimension can be compressed to 1 time-step before operating convolution. That is why S-ACE can further decrease when "W/S/T=222" and more elaborations are in the appendix for better understanding. 
In conclusion, by comparing with those advanced methods on CIFAR-10 and CIFAR-100, we demonstrate significant improvements in memory efficiency, computational requirements, and storage consumption.

\subsubsection{ImageNet} We further apply our methods to SEW-ResNet34 to demonstrate superiority on the up-scaled dataset. As listed in  Table~\ref{tab:comparison on imagenet}, compared with those using improved learning techniques or special modules  \citep{rathi2020enabling, zheng2021going,li2021differentiable,hu2024advancing}, we can achieve magnitudes of memory and computation savings without using any of their tricks. 
Attention SNN \citep{yao2023attention} has comparable S-ACE (56.73 G) with ours (30.74 G and 54.69 G), but its memory efficiency, model accuracy, and storage requirement lag far behind those of our methods. We can achieve at least 1.42\% increased accuracy while reducing 45.81\% S-ACE and 83.86\% model size. Besides, our models would not require sophisticated attention modules \citep{yao2023attention} as they do, which indicates a more straightforward inference dataflow and a better computing flow mapping from software to hardware.
Furthermore, we can save 91.2\% memory (5.61  v.s. 64) and  85.83\% computation (30.74 G v.s. 217 G) while achieving a higher accuracy (70.57\%  v.s. 70.13\%) when comparing with the quantized SNN \citep{shen2024conventional}. As the bit budget increases, the model accuracy can be advanced to 72.82\% with low memory and computation overheads, outperforming full-precision SNNs  \citep{guo2024ternary, fang2021deep, deng2021temporal}. Conclusively, scaling our model to the large ImageNet-1k dataset further demonstrates the effectiveness of the proposed methods in improving the efficiency and accuracy of SNNs.

\begin{table*}[width=1.\textwidth,cols=10,pos=t]
\centering
\caption{Ablations on adaptive bit width  and the renewal mechanism. Different fonts denote the \textbf{first}, the 
{\color[HTML]{9B9B9B} \textbf{second}}, and the  \underline{third} best results, respectively. Our bit-adaptive models are initial to W/S/T=4/4/2.}
\label{exp:ablation on learnable bit}
\begin{adjustbox}{max width=\textwidth}
\begin{threeparttable}
% \begin{tabular}{c|c|cc||cc|cc|cc}
\begin{tabular}{cccccccccc}
\toprule
% \textbf{Target}         & \textbf{U-quant.} & \multicolumn{2}{c||}{\textbf{LBW}} & \multicolumn{2}{c|}{\textbf{\begin{tabular}[c]{@{}c@{}}LBW\\ +Weight-only Renewal\end{tabular}}} & \multicolumn{2}{c|}{\textbf{\begin{tabular}[c]{@{}c@{}}LBW\\ +Bilateral Renewal\end{tabular}}} & \multicolumn{2}{c}{\textbf{\begin{tabular}[c]{@{}c@{}}LBW\\ +Act.-only Renewal\end{tabular}}} \\ 
\textbf{Target}         & \textbf{U-quant.} & \multicolumn{2}{c}{\textbf{Bit-adaptive}} & \multicolumn{2}{c}{\textbf{\begin{tabular}[c]{@{}c@{}}Bit-adaptive\\ +Weight-only Renewal\end{tabular}}} & \multicolumn{2}{c}{\textbf{\begin{tabular}[c]{@{}c@{}}Bit-adaptive\\ +Bilateral Renewal\end{tabular}}} & \multicolumn{2}{c}{\textbf{\begin{tabular}[c]{@{}c@{}}Bit-adaptive\\ +Act.-only Renewal\end{tabular}}} \\ 
\midrule
\textbf{W/S/T} & \textbf{Top-1}    & \textbf{W/S/T}  & \textbf{Top-1}  & \textbf{W/S/T}                                 & \textbf{Top-1}                                & \textbf{W/S/T}                    & \textbf{Top-1}                                           & \textbf{W/S/T}                    & \textbf{Top-1}                                          \\ \midrule
1/1/1          & 74.82             & 1.60/1.91/1.0   & \underline{95.65}     & 1.73/1.98/1.0                                  & 93.28                                         & 1.25/1.36/1.0                     & {\color[HTML]{9B9B9B} \textbf{95.84}}                    & 1.38/1.32/1.0                     & {\color[HTML]{000000} \textbf{96.15}}                   \\
1/2/1          & 86.8              & 1.48/2.00/1.0   & \underline{95.55}     & 1.57/2.12/1.0                                  & 93.52                                         & 1.25/2.04/1.0                     & {\color[HTML]{9B9B9B} \textbf{96.25}}                    & 1.42/2.08/1.0                     & \textbf{96.58}                                          \\
2/2/1          & 88.65             & 1.88/2.06/1.0   & 94.62           & 1.99/2.06/1.0                                  & \underline{95.71}                                   & 2.00/1.89/1.0                     & {\color[HTML]{9B9B9B} \textbf{96.47}}                    & 2.08/2.08/1.0                     & \textbf{96.51}                                          \\
2/3/1          & 93.64             & 2.10/3.00/1.0   & 96.37           & 1.96/3.00/1.0                                  & \underline{96.51}                                   & 2.02/3.06/1.0                     &\textbf{96.97}                   & 2.24/3.06/1.0                     & {\color[HTML]{9B9B9B} \textbf{96.63}}                   \\
3/3/1          & 95.38             & 3.01/2.95/1.0   & 96.34           & 3.00/3.06/1.0                                  & \underline{96.66}                                   & 2.87/3.04/1.0                     & {\color[HTML]{9B9B9B} \textbf{96.77}}                    & 2.89/2.87/1.0                     & \textbf{96.90}                                          \\
3/4/1          & 96.09             & 2.87/4.10/1.0   & 96.72           & 2.90/3.98/1.0                                  & \underline{96.89}                                   & 2.97/3.94/1.0                     & {\color[HTML]{9B9B9B} \textbf{97.01}}                    & 3.01/4.00/1.0                     & \textbf{97.02}                                          \\
4/4/1          & 96.35             & 4.04/4.06/1.0   & 96.78           & 3.99/4.02/1.0                                  & \underline{97.00}                                   & 3.86/4.00/1.0                     & {\color[HTML]{9B9B9B} \textbf{97.03}}                    & 3.99/4.06/1.0                     & \textbf{97.05}                                          \\
4/4/2          & 96.39             & 4.02/3.93/1.92  & \underline{96.80}     & 3.88/4.04/2.0                                  & 96.57                                         & 4.01/3.97/2.0                     &   \textbf{97.03}                  & 3.93/4.03/2.08                    & {\color[HTML]{9B9B9B} \textbf{96.92}}                   \\ \bottomrule
\end{tabular}
% \begin{tablenotes}
% \footnotesize
% \item $\dagger$ denotes plain uniform quantization, following  \citep{shen2024conventional}.
% \item $*$ denotes the temporal squeezing is applied in inference.
% \end{tablenotes}
\end{threeparttable}
\end{adjustbox}
\end{table*}

\subsection{Ablation study}
\label{sbsec:ablation}
We first continue to use ResNet20 and CIFAR10 to conduct ablation studies on the proposed techniques. Then, we migrate the proposed method to a different architecture, Spikformer \citep{zhou2022spikformer}, to further verify  effectiveness. On the extension to the event dataset, we leverage our methods to mitigate the "temporal constraint" issue, which troubles the prior art \citep{shen2024conventional}.
Furthermore, we provide visualizations of the learned bit allocation as empirical evidence and statistically analyze the results.
Finally, we compare our bit-adaptive SNNs with the ANN baselines to show the superiority of  SNNs in scenarios of ultra-low-energy computation.
Notably, since our work targets building efficient and accurate SNNs, we mainly focus on the effectiveness of the proposed approaches under the low bit budget settings and conduct ablation studies in this background.

\begin{table*}[width=\textwidth,cols=10,pos=t]
\centering
\caption{Ablations on the refined spiking neuron. Different fonts denote the \textbf{first}, the 
{\color[HTML]{9B9B9B} \textbf{second}}, and the  \underline{third} best results, respectively.}
\label{exp:ablation on refined neuron}
\begin{adjustbox}{max width=\textwidth}
\begin{threeparttable}
% \begin{tabular}{c|c|cc|cc|cc|cc}
% \toprule
% {  \textbf{Target} }                                                                                  & {  \textbf{\begin{tabular}[c]{@{}c@{}}$\tau=2$\& Equation~\ref{eq:mlif3} \\ U-quant.\end{tabular}}} & \multicolumn{2}{c|}{{  \textbf{\begin{tabular}[c]{@{}c@{}}$\tau=2$\& Equation~\ref{eq:mlif3}  \\ w/o renewal\end{tabular}}}} & \multicolumn{2}{c|}{{  \textbf{\begin{tabular}[c]{@{}c@{}}$\tau=2$\& Equation~\ref{eq:mlif3} \\ w/ renewal\end{tabular}}}} & \multicolumn{2}{c|}{{  \textbf{\begin{tabular}[c]{@{}c@{}}$\tau=2$\& Equation~\ref{eq:our sn3}\\ w/ renewal\end{tabular}}}} & \multicolumn{2}{c}{{  \textbf{\begin{tabular}[c]{@{}c@{}}$\tau=1$\& Equation~\ref{eq:our sn3}\\ w/ renewal\end{tabular}}}} \\ 
\begin{tabular}{cccccccccc}
\toprule
{\textbf{Target} }                                                                                  & {\textbf{\begin{tabular}[c]{@{}c@{}}$\tau=2$ \& Flooring \\ U-quant.\end{tabular}}} & \multicolumn{2}{c}{{\textbf{\begin{tabular}[c]{@{}c@{}}$\tau=2$ \& Flooring  \\ w/o renewal\end{tabular}}}} & \multicolumn{2}{c}{{\textbf{\begin{tabular}[c]{@{}c@{}}$\tau=2$ \& Flooring \\ w/ renewal\end{tabular}}}} & \multicolumn{2}{c}{{\textbf{\begin{tabular}[c]{@{}c@{}}$\tau=2$ \& Rounding\\ w/ renewal\end{tabular}}}} & \multicolumn{2}{c}{{\textbf{\begin{tabular}[c]{@{}c@{}}$\tau=1$ \& Rounding\\ w/ renewal\end{tabular}}}} \\

\midrule
{\textbf{W/S/T}} &   \textbf{Top-1}                                                       &   \textbf{W/S/T}                    &   \textbf{Top-1}                  &   \textbf{W/S/T}                   &  \textbf{Top-1}                 &   \textbf{W/S/T}                   &  \textbf{Top-1}                  &   \textbf{W/S/T}                  &  \textbf{Top-1}                  \\ \midrule
  1/1/1                                                                              &  74.82                                                                & 1.80/2.14/1.0                    &   95.12                            &   1.80/2.09/1.0                    & \underline{95.58}                     &   1.46/1.34/1.0                    & {\color[HTML]{9B9B9B} \textbf{95.71}}                  &   1.38/1.32/1.0                   &   \textbf{96.15}                  \\
  1/2/1                                                                              &   10.02                                                                &   1.77/2.2/1.0                      &   95.21                            &   1.58/2.13/1.0                   & \underline{96.14}                     &   1.40/2.05/1.0                   & {\color[HTML]{9B9B9B} \textbf{96.48}}                  &   1.42/2.08/1.0                   &   \textbf{96.58}                  \\
  2/2/1                                                                             &   0.1                                                                  &   2.10/2.07/1.0                     &   95.50                            &   2.09/1.97/1.0                    &   \underline{96.04}                    &   1.88/1.95/1.0                    & {\color[HTML]{9B9B9B} \textbf{96.45}}                  &  2.08/2.08/1.0                   &   \textbf{96.51}                  \\
  2/3/1                                                                              &   89.06                                                                &   2.05/3.01/1.0                     & \underline{96.51}                      &   2.13/3.16/1.0                    &   96.38                           &   1.96/3.04/1.0                    &   \textbf{96.72}                  &   2.24/3.06/1.0                   & {\color[HTML]{9B9B9B} \textbf{96.63}}                  \\
  3/3/1                                                                              &   0.1                                                                  &   2.98/3.02/1.0                     &   96.18                           &   3.01/3.07/1.0                    & \underline{96.43}                     &   3.03/3.01/1.0                    & {\color[HTML]{9B9B9B} \textbf{96.82}}                  & 2.89/2.87/1.0                   &   \textbf{96.90}                 \\
3/4/1                                                                              &   96.31                                                                &   3.06/3.98/1.0                     &   96.64                            &   2.99/4.02/1.0                    & \underline{96.80}                     &   2.95/4.06/1.0                    & {\color[HTML]{9B9B9B} \textbf{97.00}}                  &   3.01/4.0/1.0                    &   \textbf{97.02}                  \\ \bottomrule
\end{tabular}

\end{threeparttable}
\end{adjustbox}
\end{table*}

\subsubsection{Effectiveness of adaptive bit width and the renewal mechanism}
\label{subsec:ablation on bie learn and renewal}
We first perform the plain uniform quantization on SNNs (U-quant.) as the baselines, following  \citep{shen2024conventional}. As shown in Table~\ref{exp:ablation on learnable bit}, the proposed adaptive bit-width technology consistently outperforms U-quant., managing to allocate valid bit widths to different layers and improving the model accuracy to a higher level. Furthermore, we add the proposed renewal mechanism only to the spike activation ($V_{th,l}^1$, Act.-only), or only to the weight ($S_q^l$, Weight-only), or to the bilateral case ($V_{th,l}^1$ and $S_q^l$). The results coherently show the renewal mechanism can further improve the outcomes of the bit-adaptive models, mitigating the step-size mismatch issue.
As described before, SNN's activation would accumulate quantization errors along the temporal dimension. That is why Act.-only renewal is effective enough to produce the best results. Bilateral renewal seconds due to  a slightly lower bit budget. For simplicity, we adopt the 
Act.-only renewal as the main method by default.

{ In addition, another interesting finding is that the results of the target W/S/T=4/4/1 are better than those of the target W/S/T=4/4/2. This may imply that more repeated static information  contribute limited improvement or even cause over-fitting to the processing of static data when spatial information is efficiently provided and processed.} 

\begin{table*}[width=.7\textwidth,cols=8,pos=t]
\centering
\caption{Ablations of Spikformer-4-384 on CIFAR10. Different fonts denote the \textbf{first} and the 
{\color[HTML]{9B9B9B} \textbf{second}} best results, respectively.}
\label{exp:ablation of spf}
\begin{adjustbox}{max width=0.7\textwidth}
\begin{threeparttable}
% \begin{tabular}{c|c|cc|cc|cc}
% \toprule
% { \textbf{ Target }}                                                                                   & {  \textbf{U-quant.}} & \multicolumn{2}{c|}{{  \textbf{LBW}}}                & \multicolumn{2}{c|}{{  \textbf{LBW + renewal}}}                 & \multicolumn{2}{c}{\textbf{Origin \citep{zhou2022spikformer}}}          \\ \midrule
% {  \textbf{W/S/T}} & {  \textbf{Top-1}}    & {  \textbf{W/S/T}} & {  \textbf{Top-1}} & {  \textbf{W/S/T}} & {  \textbf{Top-1}} & \textbf{W/S/T}           & Top-1                   \\ 
\begin{tabular}{cccccccc}
\toprule
{\textbf{Target}}                                                                                   & {  \textbf{U-quant.}} & \multicolumn{2}{c}{{\textbf{Bit-adaptive}}}                & \multicolumn{2}{c}{{\textbf{Bit-adaptive + renewal}}}                 & \multicolumn{2}{c}{\textbf{Origin \citep{zhou2022spikformer}}}          \\ \midrule
{\textbf{W/S/T}} & {\textbf{Top-1}}    & {\textbf{W/S/T}} & {\textbf{Top-1}} & {\textbf{W/S/T}} & {\textbf{Top-1}} & \textbf{W/S/T}           & \textbf{Top-1}                   \\ 

\midrule
{  1/1/1}                                                                              & {  79.53}             & {  1.44/1.48/1.0}  & {\color[HTML]{9B9B9B} \textbf{95.37}} & {  1.53/1.5/1.0}   & {  \textbf{95.47}} &                          &                         \\
{  1/2/1}                                                                              & {  89.3}              & {  1.47/1.99/1.0}  & {\textbf{95.68}} & {  1.47/2.01/1.0}  & {  \color[HTML]{9B9B9B} \textbf{95.54}} &                          &                         \\
{  2/2/1}                                                                              & {  89.68}             & {  1.88/2.03/1.0}  & {\color[HTML]{9B9B9B} \textbf{95.80}} & {  2.02/2.01/1.0}  & {  \textbf{95.84}} &                          &                         \\
{  2/3/1}                                                                              & {  94.11}             & {  2.00/2.94/1.0}  & {\color[HTML]{9B9B9B} \textbf{95.93}} & {  2.15/2.95/1.0}  & {  \textbf{96.04}} &                          &                         \\
{  3/3/1}                                                                              & {  94.21}             & {  2.93/3.01/1.0}  & {\color[HTML]{9B9B9B} \textbf{95.87}} & {  2.89/2.99/1.0}  & {  \textbf{95.99}} & \multirow{-5}{*}{16/1/4} & \multirow{-5}{*}{95.19} \\ \bottomrule
\end{tabular}
\begin{tablenotes}
\footnotesize
\item U-quant. denotes plain uniform quantization, following  \citep{shen2024conventional}.
\end{tablenotes}
\end{threeparttable}
\end{adjustbox}
\end{table*}

\subsubsection{Importance of the refined spiking neuron}
We further verify the effectiveness of the refined spiking neuron. Key improvements over prior multi-bit spiking neurons \citep{guo2024ternary,xiao2024multi} include:  \textbf{(1)} refining  Equation~\ref{eq:mlif3} to  Equation~\ref{eq:our sn3} by replacing flooring with rounding, and \textbf{(2)} adopting the IF neuron ($\tau=1$) instead of the default LIF neuron ($\tau=2$) \citep{fang2023spikingjelly}. In  Table~\ref{exp:ablation on refined neuron}, the results consistently show that the original multi-bit neuron would impair the quantized model's robustness, whereas our refinement method and renewal mechanism steadily improve the model performance to better states. This enables the development of high-performance and low-overhead SNNs.

Another insight might also be drawn. In the field of deep learning, the value quantizer regulates information widths of features and consequently reduce the redundant data movement and computing operations in the following layer. 
The firing mechanism in the LIF model emulates biological neurotransmission by transforming chemical stimuli into electrical impulses, achieving energy-efficient information transfer \citep{gerstner2014neuronal}. Therefore, the firing mechanism is the key to generating the efficient value representation, i.e., low-bit value, to achieve energy-efficient computing as the value quantizer does. Previous works \citep{wu2019direct,hu2021spiking,guo2022real} on direct-trained SNNs directly adopted the mathematical formulation of the LIF model without adapting it to the intrinsic characteristics of deep learning computations, which potentially compromises the network performance. After introducing the concept of quantization error and adhering to the  principle of minimizing quantization errors, we refined the LIF model, thereby achieving further performance enhancements in SNNs. In conclusion, although SNN is a bio-inspired model, reducing the quantization error is still important in maintaining SNN's good model performance.

\begin{figure*}[b]
  \centering
  \includegraphics[width= .8\textwidth]{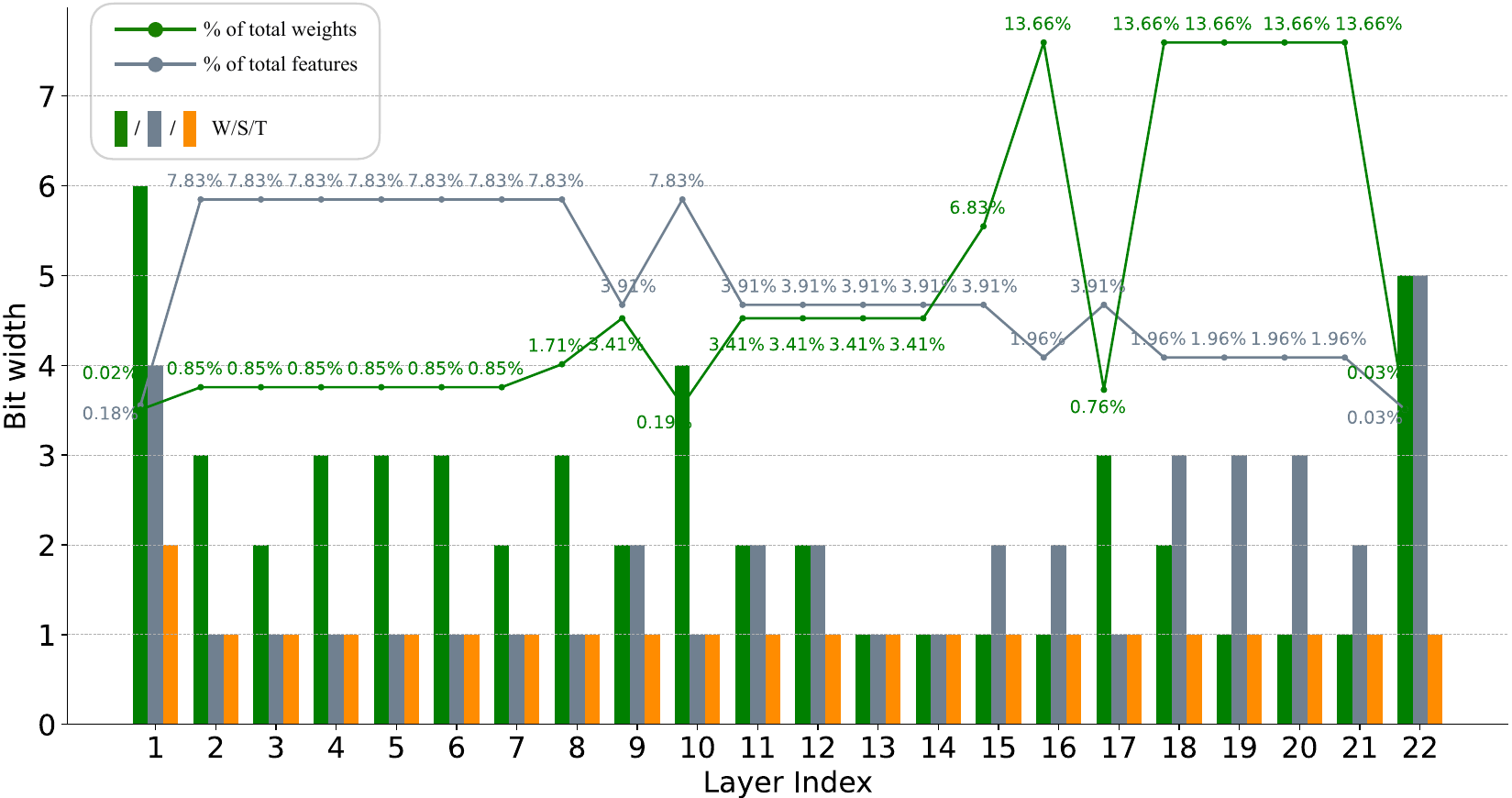}
  \caption{Statistics of each layer's bit widths. Since layer \#1 has two time-steps, we report the averaged spike bit width for clarity.}
  \label{fig:bit_vis}
\end{figure*}

\subsubsection{Validation on spiking self-attention architectures}
Migrating our methods to Spikformer is non-trivial because the self-attention operation is a different computing paradigm where a new quantizable tensor $atten$ (see Figure~\ref{fig:overview}(a)) is produced. We would like to leave room for future work. 
Here, we naively apply our methods to Spikformer as depicted in  Figure~\ref{fig:overview} only to validate the migratability and effecitveness of the proposed methods, and avoid competing with those state-of-the-art works that are based on sophisticated spiking self-attention architectures. 
We adopt the default Spikformer-4-384 \citep{zhou2022spikformer} on CIFAR10. In Table~\ref{exp:ablation of spf}, our bit-adaptive model surpasses the original Spikformer, while plain uniform quantization fails. Besides, adding the renewal mechanism further lifts accuracy.

\subsubsection{Mitigating the temporal constraint}
We further test the migratability of the proposed methods on the event datasets. We use the same default Spikformer and settings as  \citep{shen2024conventional} without data augmentation. 
As shown in  Table~\ref{exp:ablations on dvs}, "temporal constraint" refers to the dramatic accuracy decrease of event datasets as temporal length declines to the extremely low time-step \citep{shen2024conventional}. However, in our work, such a phenomenon disappears and accuracy is well maintained even in the case of $T=1$. 
This implies that our methods can better extract spatial information even in sparse event data.
Meanwhile, we are in line with prior findings that temporal length would maintain a notable effect on accuracy for event datasets. 
Moreover, compared with the original Spkiformer, our methods can achieve higher accuracy with lower bit budgets
and time-steps, proving the effectiveness on event data.

Futhermore, another mainstream event-based dataset DVS-GESTURE \citep{amir2017low} is smaller and easier for models to converge, and the  "temporal constraint" issue does not manifest. In this case, our model can still significantly strengthen the temporal processing ability of SNNs. As listed in Table~\ref{exp:ablations on dvsg}, our model can achieve consistently better accuracy while using shorter time lengths and fewer bit budgets when compared with the baseline models. In the case of W/S/T = 1.17/2.18/6.0, our model can achieve advanced 99.31 \% top-1 accuracy. To the best of our knowledge, this should be a state-of-the-art result of SNNs on the dynamic dataset.

\begin{table}[width=\linewidth,cols=6,pos=t]
\centering
\caption{Spikformers on CIFAR10-DVS. The \textbf{best results} and \underline{temporal lengths} are highlighted.}
\label{exp:ablations on dvs}
\begin{adjustbox}{max width=\linewidth}
\begin{threeparttable}
% \begin{tabular}{cc|cc|cc}
% \toprule
% \multicolumn{2}{c|}{\textbf{Origin} \citep{shen2024conventional}}             & \multicolumn{2}{c|}{\textbf{Quant. Spkiformer} \citep{shen2024conventional}} & \multicolumn{2}{c}{\textbf{Our Spkiformer}} \\ 
\begin{tabular}{cccccc}
\toprule
\multicolumn{2}{c}{\textbf{Origin} \citep{shen2024conventional}}             & \multicolumn{2}{c}{\textbf{Quantized Spkiformer} \citep{shen2024conventional}} & \multicolumn{2}{c}{\textbf{Our Spkiformer}} \\ 
\midrule
\textbf{W/S/T}           & \textbf{Top-1}        & \textbf{W/S/T}      & \textbf{Top-1}      & \textbf{W/S/T}    & \textbf{Top-1}    \\ \midrule
\multirow{5}{*}{16/1/16} & \multirow{5}{*}{80.7} & 1/1/\underline{16}             & 79.8       & 1.11/2.00/\underline{6.0}     & \textbf{80.9}     \\
                         &                       & 1/2/\underline{8}             & 79.3                & 1.28/2.00/\underline{4.0}     & \textbf{80.0}              \\
                         &                       & 1/4/\underline{4}            & 63.1                & 1.33/4.00/\underline{3.0}     & \textbf{80.1}              \\
                         &                       & 1/8/\underline{2}            & 43.0                & 1.31/4.03/\underline{2.0}     & \textbf{78.7}              \\
                         &                       & 1/16/\underline{1}          & 35.8                & 1.31/4.03/\underline{1.0}     & \textbf{77.4}              \\ \bottomrule
\end{tabular}
% \begin{tablenotes}
% \footnotesize
% \item \textbf{TS} abbreviates temporal squeezing.
% \end{tablenotes}
\end{threeparttable}
\end{adjustbox}
\end{table}

\begin{table}[width=\linewidth,cols=6,pos=t]
\centering
\caption{Spikformers on DVS-GESTURE. The \textbf{best results} and \underline{temporal lengths} are highlighted.}
\label{exp:ablations on dvsg}
\begin{adjustbox}{max width=\linewidth}
\begin{threeparttable}
% \begin{tabular}{cc|cc|cc}
% \toprule
% \multicolumn{2}{c|}{\textbf{Origin} \citep{shen2024conventional}}             & \multicolumn{2}{c|}{\textbf{Quant. Spkiformer} \citep{shen2024conventional}} & \multicolumn{2}{c}{\textbf{Our Spkiformer}} \\ 
\begin{tabular}{cccccc}
\toprule
\multicolumn{2}{c}{\textbf{Origin \citep{shen2024conventional,zhou2022spikformer}}}              & \multicolumn{2}{c}{\textbf{Quantized Spkiformer \citep{shen2024conventional}}} & \multicolumn{2}{c}{\textbf{Our Spkiformer}} \\ \midrule
\textbf{W/S/T}           & \textbf{Top-1}        & \textbf{W/S/T}         & \textbf{Top-1}        & \textbf{W/S/T}       & \textbf{Top-1}       \\ \midrule
\multirow{5}{*}{16/1/16} & \multirow{5}{*}{98.3} & 1/1/\underline{16}                 & 96.67                 & 1.17/2.18/\underline{6.0}          &\textbf{99.31}       \\
                         &                       & 1/2/\underline{8}                  & 98.48       & 1.29/2.08/\underline{4.0}          & \textbf{98.61}                \\
                         &                       & 1/4/\underline{4}                  & 97.35                 & 1.32/4.00/\underline{3.0}          & \textbf{97.92}                \\
                         &                       & 1/8/\underline{2}                  & 96.59                 & 1.31/4.00/\underline{2.0}          & \textbf{97.57}                \\
                         &                       & 1/16/\underline{1}                 & 95.45                 & 1.37/4.00/\underline{1.0}          & \textbf{96.18}                \\ \bottomrule
\end{tabular}
% \begin{tablenotes}
% \footnotesize
% \item \textbf{TS} abbreviates temporal squeezing.
% \end{tablenotes}
\end{threeparttable}
\end{adjustbox}
\end{table}

\subsubsection{Statistics of bit widths} 
We visualize the learned bit allocation of ResNet20 with W/S/T = 1.38/1.32/1 on CIFAR10 to visually confirm the effectiveness. Notably, different from any prior arts \citep{guo2024ternary,zheng2021going,fang2021deep}, we add additional spiking neurons before the first layer to fully benefit the model with addition-only computation. In consistence with the previous setting, the model is initialized to W/S/T = 4/4/2. 
As vividly displayed in  Figure~\ref{fig:bit_vis}, \emph{\textbf{I)}} the layers with the higher proportion of features and weights tend to get lower bits thus lowering the overall memory and computation. \emph{\textbf{II)}} The first and last layers tend to learn higher bit widths in line with their outstanding importance as reported in prior arts \citep{choi2018pact}. With \emph{\textbf{I)}} and \emph{\textbf{II)}} combined, we conclude our adaptive bit allocation method should be sensible and practical. 

\subsubsection{Additional comparisons with mixed-precision ANNs} \label{sbsec:compare with anns}
Since we use a similar bit-width parametrization and gradient calculation to the mixed-precision ANN \citep{uhlich2019mixed}, we also need to conduct a fair comparison with mixed-precision ANNs. Here, we continue using ResNet20 on CIFAR10.

\begin{table}[width=\linewidth,cols=5,pos=t]
\centering
\caption{Additional comparisons with mixed-precision ANNs.}
\label{exp:comparison with mix ann}
\begin{adjustbox}{max width=\linewidth}
\begin{threeparttable}
\begin{tabular}{ccccc}
\toprule
\textbf{Target}   & \multicolumn{2}{c}{\textbf{Mixed-precision ANN} \citep{uhlich2019mixed}} & \multicolumn{2}{c}{\textbf{Our bit-adaptive SNN}} \\ \midrule
\textbf{W/A(S)/T} & \textbf{W/A}           & \textbf{Top-1}           & \textbf{W/S/T}                                     & \textbf{Top-1}                                     \\ \midrule
1/1/1             & 1.48/1.93              & 95.78                    & 1.25/1.36/1.0                                      & \textbf{95.84}                                     \\
1/2/1             & 1.48/2.03              & 95.68                    & 1.25/2.04/1.0                                      & \textbf{96.25}                                     \\
2/2/1             & 1.92/2.07              & 95.71                    & 2.00/1.89/1.0                                      & \textbf{96.47}                                     \\
2/3/1             & 2.19/3.05              & 96.29                    & 2.02/3.06/1.0                                      & \textbf{96.97}                                     \\
3/3/1             & 3.13/2.97              & 96.56                    & 2.87/3.04/1.0                                      & \textbf{96.77}                                     \\
3/4/1             & 3.07/3.98              & 96.97                    & 2.97/3.94/1.0                                      & \textbf{97.01}                                     \\
4/4/1             & 3.87/3.96              & 96.76                    & 3.86/4.00/1.0                                      & \textbf{97.03}                                     \\ \bottomrule
\end{tabular}
\begin{tablenotes}
\footnotesize
\item The results of mixed-precision ANNs \citep{uhlich2019mixed} are produced by re-implementing their paper.
\end{tablenotes}
\end{threeparttable}
\end{adjustbox}
\end{table}

As listed in Table~\ref{exp:comparison with mix ann}, our proposed adaptive bit allocation SNNs consistently outperform the mixed-precision ANNs under every W/S/T setting. In the case of extremely low bit widths, e.g., the target W/S/T=1/1/1, our SNN can achieve higher accuracy while using a lower bit budget. This implies that SNNs may have an advantage over ANNs in scenarios of ultra-low-energy computation, which is in line with previous studies \citep{deng2020rethinking,he2020comparing}. 

{Additionally, our SNN has adopted the proposed renewal mechanism and a bit different gradient calculations (see Section~\ref{sbsec:grad cal}). Therefore, Table~\ref{exp:comparison with mix ann} can also imply the effectiveness of our proposed methods as a whole.}

% \begin{table}[t]
% \centering
% \caption{Comparisons with existing works on ImageNet.}
% \label{exp:comparison with ann baseline}
% \begin{adjustbox}{max width=0.48\textwidth}
% \begin{threeparttable}

% % \begin{tablenotes}
% % \footnotesize
% % \item $\dagger$ denotes plain uniform quantization, following  \citep{shen2024conventional}.
% % \item $*$ denotes the temporal squeezing is applied in inference.
% % \end{tablenotes}
% \end{threeparttable}
% \end{adjustbox}
% \end{table}

\section{Discussion}
\label{sec:discussion}

\subsection{Maintaining the neuromorphic computing nature}
The most precious characteristic that SNN brings is its asynchronous processing ability, i.e., event-driven and clock-free computing. We would like to address the potential concerns about how such nature and merit could be maintained when temporal squeezing is applied.

\textbf{Exploiting the spike perceptrons.}  {Temporal squeezing itself is consistent with neurodynamics as it is essentially composed of a signal accumulation \citep{gerstner2014neuronal} and a division operation.} A similar accumulation mechanism can also be found in the current accumulation stage of spiking neuron, such as LIF and IF. Particularly, SpikeConverter neuron \citep{liu2022spikeconverter} separates this accumulation from the spiking neuron modeling and uses a special computing pipeline to handle this accumulation. Their accumulation is identical to ours, which is proven feasible in asynchronous computing. In addition,
such accumulation has also been realized and implemented
by neuromorphic circuits to process the DVS signals and
generate the dynamic datasets \citep{amir2017low,lin2021imagenet}. 

\textbf{Using the two-stage pipeline.} Although we have theoretically elaborated on the harmlessness of temporal squeezing to the asynchronous nature in the above, one may argue that the entire network's spike flow could be interrupted by the squeezing stage, leading to longer inference latency. This concern has already been discussed and resolved by \citep{liu2022spikeconverter}, where the two-stage pipeline is adopted. The first stage is spike accumulation, i.e., the temporal squeezing, the second stage is spike emission. These two separate stages can overlapped inter-layer. Therefore, using temporal squeezing only introduces very limited latency increase. 

{
\subsection{Hardware considered energy merit analysis}
\label{sbsec:energy merits over ann}
\citep{davidson2021comparison} explains that SNNs are not inherently more energy-efficient than ANNs in all scenarios, detailing a
specific condition where SNN energy becomes lower. This specific condition demands "the expected number of spikes per output" to be less than 1.72 on their hardware implementation. 

As shown in Table~\ref{tab:comparison on cifar} and Table~\ref{tab:comparison on imagenet}, our the temporal lengths of our SNNs are mostly learned to 1.
Clearly, "the expected number of spikes per output" is less than 1.72. However, we need to consider a more rigorous situation where "the expected number of spikes per output" is actually "the expected number of non-zero bits in activation states per output", because our spiking neuron generates a multi-bit spike at a time-step while \citep{davidson2021comparison} generates a 1-bit spike at a time-step. Thus, we can have:
\begin{align}
\text{activation states} &= T \cdot S,\\
\begin{split}
\text{Exp(act.)} 
&= \text{Exp} (T \cdot S \cdot fr_s)\\
&\approx T \cdot S \cdot \overline{fr}_s.
\end{split}
\end{align}
Here, Exp(act.) abbreviates "the expected number of non-zero bits in activation states per output". Given  Equation~\ref{eq:sace} and Equation~\ref{eq:nsace},  we can estimate Exp(act.) via: 
\begin{equation}
\begin{split}
\text{Exp(act.)} &\approx T \cdot S \cdot \overline{fr}_s\\
&= T\cdot S \cdot \frac{\text{NS-ACE}}{\text{S-ACE}}. 
\end{split}
\end{equation}
}
{

Given Table~\ref{tab:comparison on imagenet}, we can get our SNNs' Exp(act.)s are 0.55, 1.23, and 1.22, respectively. We also record their "average numbers of non-zero bits in activation states per output" through sampling the data from a random mini-batch of ImageNet. The recorded "average numbers of non-zero bits in activation states per output" are 
0.57, 1.28, and 1.22, respectively.

Both of the estimated and recorded results are all less than 1.72, which implies our SNNs should be more energy-efficient than ANNs on the hardware implementation of \citep{davidson2021comparison}.

\subsection{Reusable neural integration unit design}
\label{sbsec:circuits}

\begin{figure}[t]
  \centering
  \includegraphics[width= 0.5\textwidth]{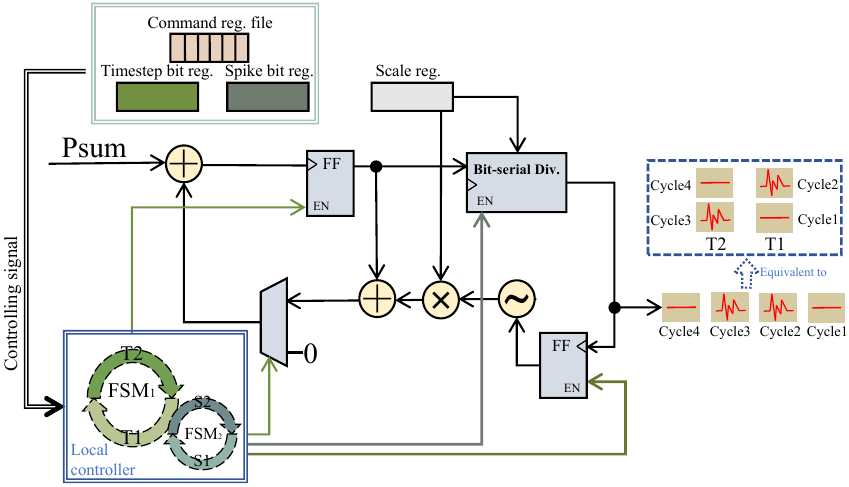}
  \caption{Bit-serial style  neural integration unit that can handle various W/S/T combinations. Psum represents the input current. "Reg." denotes register. "Div." is divider. For simplicity, we set S=2 and T=2.}
  \label{fig:mpLIF circuit}
\end{figure}

One may argue that \citep{shen2024conventional} proved the "Bit Budget" concept based on the static/unconfigurable neural integration unit. Our proposed layer-wise bit allocation would incur massive hardware resource waste if the neural integration units were dedicated to specific W/S/T combinations and could not be reused across different layers. Thus, the "Bit Budget" concept may become ineffective.

Here, we draw the design philosophy of bit-serial hardware \citep{sharma2018bit,albericio2017bit,sharify2019laconic,sharify2018loom} to provide a highly reusable neural integration unit that can handle any W/S/T combination, as shown in Figure~\ref{fig:mpLIF circuit}. For simplicity and flexibility, we utilize two counters to implement the two finite state machines (FSMs) for the timestep loop (T) and the spike bit loop (S), respectively. Thus, the FSMs become easy to configure, without any configuration overhead and resource waste. Simply adding some necessary peripheral logic circuits to generate the enabling signals will make the local controller functional. For different T and S combinations, we only need to reset the counters' count limits. Another design core is the bit-serial style divider \citep{parhami1999computer} that outputs one bit value at one cycle. Based on this, the whole unit only handles the  computation of one bit value at one cycle. \textit{In short, one bit in, and one bit out}. As for the following matrix multiplications between spikes and weights, e.g., convolutions, employing  one of those advanced mixed-precision architectures  \citep{tahmasebi2024flexibit,luo2023deepburning, li2022bit, zhang2024mixpe} would effectively and efficiently suffice.

Therefore, with our bit-serial style design, any W/S/T combination can be processed by the same neural integration unit.  The computations of different layers can be mapped to the same processing-element (PE) array. Thus, the potential hardware utilization issue can be resolved. 
}

\begin{figure}[t]
  \centering
  \includegraphics[width= 0.48\textwidth]{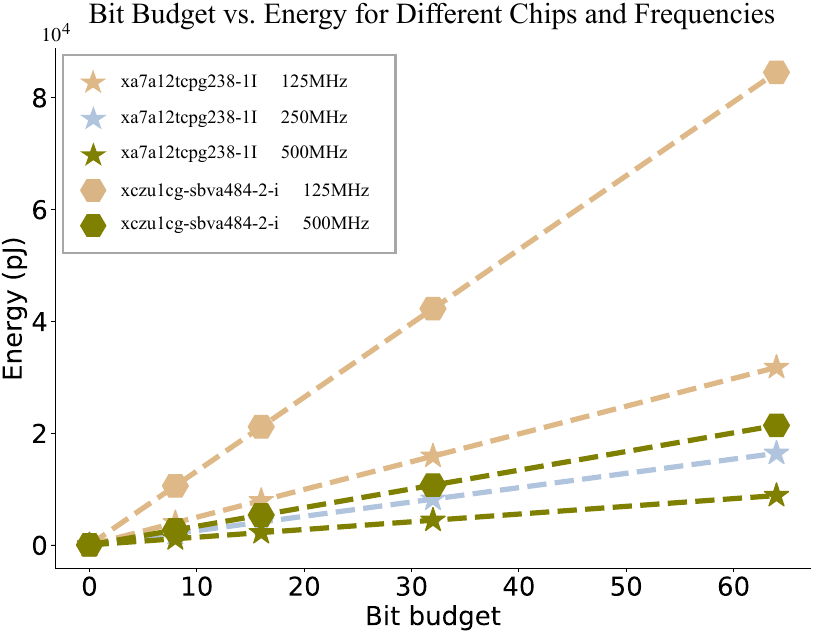}
  \caption{ Relationship between bit budgets and the energy consumption of individual synaptic operation.}
  \label{fig:sop nenergy}
\end{figure}

{
\subsection{ Energy consumption analysis of individual synaptic operation}
\label{sbsec:individual synaptic operation energy}

Based on our bit-serial style hardware implementation, we can also prove the same linear relationship between  bit budget and energy consumption for individual synaptic operations on the FPGA platform, as \citep{shen2024conventional} proved before.

As shown in Figure~\ref{fig:sop nenergy}, we conduct the analogous energy analysis as \citep{shen2024conventional}. The linear relationship between bit budgets and synaptic energy is very intuitive. Therefore, we can conclude that the "Bit Budget" concept should be consistently effective for our layer-wise bit-adaptive SNNs.
}
\section{Conclusion}
\label{sec:conclusion}
In this paper, we propose an adaptive bit allocation method, based on learnable bit widths, for efficient and accurate SNNs. 
To resolve the new challenges brought by the temporal dimension of SNN, we refine the spiking neuron to enable temporal matching inter-layer and intra-layer, and increase the adaptivity via the better neuron formulation and hyperparameter. We further formulate the step-size mismatch issue that would temporally deteriorate, and alleviate it through the proposed renewal mechanism.

Extensive experiments are conducted to demonstrate the effectiveness of the proposed methods on the static CIFAR-10, CIFAR-100, and ImageNet datasets and  the dynamic CIFAR10-DVS and DVS-GESTURE datasets. Not only is the model accuracy  improved with our approaches, but the memory (bit budget) and the computation overhead (S-ACE) are also significantly reduced. Ablation studies further prove the effectiveness of every proposed technique. Results on the dynamic datasets show our methods can further help SNNs resolve the temporal constraint issue that troubles the previous work. The effectiveness and rationality of our methods are also empirically validated through statistical visualization. Additional comparisons with the ANN baseline also imply
that SNNs may have an advantage over ANNs in scenarios
of ultra-low-energy computation. 

{
Finally, we provide some discussions and analyses concerning the potential hardware-level issues, including how to maintain the neuromorphic computing nature, hardware considered energy merit analysis, reusable neural integration unit design, and energy consumption analysis of individual synaptic operation.
 }

\section*{Acknowledgements}
This work was supported by State Grid Corporation of China, Science and Technology Project under grant No.5700-202358838A-4-3-WL.

\appendix
\section{My Appendix}

\setcounter{equation}{0}
\renewcommand{\theequation}{A.\arabic{equation}}
\setcounter{figure}{0}
\renewcommand{\thefigure}{A.\arabic{figure}}
\setcounter{table}{0}
\renewcommand{\thetable}{A.\arabic{table}}
\setcounter{algorithm}{0}
\renewcommand{\thealgorithm}{A.\arabic{algorithm}}

\subsection{Guideline}
In this technical appendix, we provide the following supplementary materials for the main text:
\begin{enumerate}
 \item Section~\ref{sec:proofs}: \textbf{(I)}  the proof of the fact that \emph{the flooring operation would incur more quantization error than the rounding operation}, \textbf{(II)} the full deduction of \textbf{Proof 1}, and \textbf{(III)} the proof of \textbf{Theorem 1} in the bidirectional domain;   
 \item Section~\ref{sec:step gradient}: the gradient calculations of $V_{th,l}^1$ and $S_q^l$; 
 \item Section~\ref{sec:2 bi gradient}: the gradient calculations of the bidirectional-spike version of the learnable bit width ($B_{s,l}$) and quantization step size ($V_{th,l}^1$);
to $V_{th,l}^1$ and $B_{s,l}$;
 \item  Section~\ref{sec: exp set}: the details of the experimental settings; 
 \item  Section~\ref{sec:elab on ts}: the elaboration on  how temporal squeezing would further reduce the S-ACE when temporal length is over 1;
 \item  Section~\ref{sec:add exps}: the additional comparisons on ImageNet and CIFAR10-DVS.
 \item  Section~\ref{sbsec:shd sec}: the potential exploration on spiking audio task.
 \item  Section~\ref{sec:maintain asyn}: more elaboration on maintaining the asynchronous nature of SNNs.
 \item  Section~\ref{sec:overall algo}: overall algorithm of the proposed adaptive-bit-allocation training.
\end{enumerate}

\subsection{Supplementary proofs}
\label{sec:proofs}
\subsubsection{Proof of the flooring operation incurring more quantization errors}
Considering the quantization process of the variable $x$, we assume $x\sim N(0,\sigma^2)$. Then, the flooring quantization process of the variable $x$ would be $x^f_q = s\cdot clip(\left\lfloor\frac{x}{s}\right\rfloor, -2^{b-1}+1, 2^{b-1}-1)$, where $s$ and $b$ are the quantization step size and the bit width, respectively. Similarly, the rounding quantization process would be $x^r_q = s\cdot clip(\left\lfloor\frac{x}{s}\right\rceil, -2^{b-1}+1, 2^{b-1}-1)$.

Defining the quantization error as $Err = (x-x_q)^2$, then we get the expectation of the flooring quantization error:
\begin{equation}
    \begin{split}
        E(Err^f) &= s^2 [ \sum_{j=0}^{2^b-3} \int_{-2^{b-1}+1+j}^{-2^{b-1}+2+j}(\frac{x}{s}-\left\lfloor\frac{x}{s}\right\rfloor)^2\cdot P(x) d\frac{x}{s}  ]\\
        &+ s^2 \int_{-\infty}^{-2^{b-1}+1}[\frac{x}{s}-(-2^{b-1}+1)]^2\cdot P(x)d\frac{x}{s}\\
        &+ s^2 \int_{2^{b-1}-1}^{+\infty}[\frac{x}{s}-(2^{b-1}-1)]^2\cdot P(x)d\frac{x}{s},\\
      \text{where } 
      P(x) &= \frac{1}{\sigma \sqrt{2\pi}} e^{-\frac{x^2}{2\sigma^2}} \text{is the PDF of }x.
    \end{split}
\end{equation}
Similarly we get the expectation of the rounding  quantization error:
\begin{equation}
    \begin{split}
        E(Err^r) &= s^2 [ \sum_{j=0}^{2^b-3} \int_{-2^{b-1}+1+j}^{-2^{b-1}+2+j}(\frac{x}{s}-\left\lfloor\frac{x}{s}\right\rceil)^2\cdot P(x) d\frac{x}{s}  ]\\
        &+ s^2 \int_{-\infty}^{-2^{b-1}+1}[\frac{x}{s}-(-2^{b-1}+1)]^2\cdot P(x)d\frac{x}{s}\\
        &+ s^2 \int_{2^{b-1}-1}^{+\infty}[\frac{x}{s}-(2^{b-1}-1)]^2\cdot P(x)d\frac{x}{s}.\\
    \end{split}
\end{equation}
For clarity, We simplify the mathematical notation in the following by omitting the upper and lower bounds of the integral symbol. Thus, we get
\begin{equation}
    \begin{split}
        &E(Err^f) - E(Err^r)\\ =& s^2 \{ \sum_j \int [(\frac{x}{s}-\left\lfloor\frac{x}{s}\right\rfloor)^2 - (\frac{x}{s}-\left\lfloor\frac{x}{s}\right\rceil)^2 ]\cdot P(x) d\frac{x}{s}  \}\\
        =&  s^2 \{ \sum_j \int \overbrace{ [\frac{2x}{s}-(\left\lfloor\frac{x}{s}\right\rfloor+\left\lfloor\frac{x}{s}\right\rceil)] [\left\lfloor\frac{x}{s}\right\rceil-\left\lfloor\frac{x}{s}\right\rfloor ]}^\#\cdot P(x) d\frac{x}{s}  
    \end{split}
\end{equation}
We further discuss $\#$ as follows:
\begin{enumerate}
    \item If $x>0$ and $\frac{x}{s}\notin \{ \mathbb {Z},  \mathbb {Z}-\frac{1}{2}\}$, then  $[\frac{2x}{s}-(\left\lfloor\frac{x}{s}\right\rfloor+\left\lfloor\frac{x}{s}\right\rceil)] > 0$ and $[\left\lfloor\frac{x}{s}\right\rceil-\left\lfloor\frac{x}{s}\right\rfloor ] > 0$. Thus, $\# > 0$.
    \item If $x<0$ and $\frac{x}{s}\notin \{ \mathbb {Z},  \mathbb {Z}-\frac{1}{2}\}$, then  $[\frac{2x}{s}-(\left\lfloor\frac{x}{s}\right\rfloor+\left\lfloor\frac{x}{s}\right\rceil)] < 0$ and $[\left\lfloor\frac{x}{s}\right\rceil-\left\lfloor\frac{x}{s}\right\rfloor ] < 0$. Thus, $\# > 0$.
    \item $\frac{x}{s}\in \{ \mathbb {Z},  \mathbb {Z}-\frac{1}{2}\}$, then $[\frac{2x}{s}-(\left\lfloor\frac{x}{s}\right\rfloor+\left\lfloor\frac{x}{s}\right\rceil)] = 0$ and $[\left\lfloor\frac{x}{s}\right\rceil-\left\lfloor\frac{x}{s}\right\rfloor ] > 0$. Thus, $\# = 0$.
\end{enumerate}
Therefore, if $x\in \{ \mathbb {Z},  \mathbb {Z}-\frac{1}{2}\}$, $\# = 0$. Otherwise, $\# > 0$. Plus, $P(x)>0$. We can get
\begin{equation}
    \begin{split}
        E(Err^f) - E(Err^r)>0  \\
        E(Err^f) > E(Err^r)
    \end{split}
\end{equation}
In conclusion, the flooring quantization would incur higher quantization error than the rounding quantization.
\subsubsection{Complete deduction of \textbf{Proof 1}}
We first recall \textbf{Proof 1}:
\\\textbf{Proof 1}\quad
Given $b'<b$ and $x'_q=s\cdot  clip(\left\lfloor\frac{x}{s}\right\rceil,0,2^{b'}-1)$, we obtain $x'_q = x_q - \lambda$, where  $\lambda=s[clip(\left\lfloor\frac{x}{s}\right\rceil,2^{b'}-1,2^b-1)-(2^{b'}-1)]$. Thus, $Err'=(x-x_q+\lambda)^2$. And, the possibility of $P(Err'>Err)$ equates to $P(\lambda > 0)$.

To prove $P(Err'>Err)$ equates to $P(\lambda > 0)$, we first discuss when $ Err'>Err$. We get
\begin{equation}
    \begin{split}
        Err'- Err &= (x-x_q)^2 + 2\lambda(x-x_q) + \lambda^2 - (x-x_q)^2\\
        &= 2\lambda(x-x_q) + \lambda^2\\
        &= \lambda(2x-2x_q+\lambda)\\
        &= \lambda(2x-x_q-(x_q-\lambda))\\
        &= \lambda(2x-x_q-x'_q)\\
    \end{split}
\end{equation}
We further discuss
\begin{enumerate}
    \item If $ \frac{x}{s} \le 2^{b'}-1$, then $\lambda = 0$. Thus, $Err'- Err = 0$.
    \item If $\frac{x}{s} = 2^{b'}-\frac{1}{2}$ 
    , then $x'_q = s(2^{b'}-1)$ and $x_q = s\cdot2^{b'}$. Thus, $(2x-x_q-x'_q)=0$. And, $Err'- Err = 0$.
    \item If $ \frac{x}{s} > 2^{b'}-1$ and $\frac{x}{s} \ne 2^{b'}-\frac{1}{2}$, then  $x'_q = s(2^{b'}-1)$ and $\lambda>0$. Thus,
\begin{enumerate}
    \item If $ 2^{b'}-1 < \frac{x}{s} < 2^{b'}-\frac{1}{2}$,  then $x'_q=x_q=2^{b'}-1<x$. Therefore, $(2x-x_q-x'_q)>0$.
    \item If $ 2^{b'}-\frac{1}{2} < \frac{x}{s} \le 2^{b'}$,  then $x-s(2^{b'}-1)>s\cdot2^{b'}-x$. Therefore, $x-x'_q>x_q-x \Rightarrow (2x-x_q-x'_q)>0$.
    \item If $2^{b'}<\frac{x}{s}$, then $x-s(2^{b'}-1)>s>\cdot2^{b'}-x$. Therefore, $(2x-x_q-x'_q)>0$.
\end{enumerate}
    In anyway,  $(2x-x_q-x'_q)>0$ and $\lambda>0$. Thus, $Err'- Err > 0$.
\end{enumerate}
In all, if $\lambda>0$ and $\frac{x}{s} \ne 2^{b'}-\frac{1}{2}$, $Err'- Err > 0$. Otherwise, $Err'- Err =0$. To simplify the equation and make the lower bound calculation, i.e. \textbf{Lemma 1}, easier, we can statistically approximate that $P(Err'>Err)$ equates to $P(\lambda > 0)$. The corresponding approximation loss is  statistically negligible as well.
\subsubsection{Theorem 1 in the bidirectional domain}
Since we have already proven \textbf{Theorem 1} when $x\sim 2N(0,\sigma^2)|x>0$ in the main text. Proving \textbf{Theorem 1} is still tenable in the bidirectional domain, i.e. $x\sim N(0,\sigma)$, is similar and easy.

First, we discuss when $x\sim N(0,\sigma^2)|x>0$. With the exact same deduction process of \textbf{Proof 1}, we get  "\emph{the quantization error would increase by the possibility of $P(x>\frac{2^{b'}-1}{2^{b}-1}\cdot3\sigma| (x\sim N(0,\sigma^2),  x>0)$}".

Second, we discuss when $x\sim N(0,\sigma^2)|x<0$. Same to the above, we get "\emph{the quantization error would increase by the possibility of $P(x<-\frac{2^{b'}-1}{2^{b}-1}\cdot3\sigma| (x\sim N(0,\sigma^2),  x<0)$}".

Since $x\sim N(0,\sigma^2)$, $P(x<-\frac{2^{b'}-1}{2^{b}-1}\cdot3\sigma| (x\sim N(0,\sigma^2),  x<0)$ would equate to $P(x>\frac{2^{b'}-1}{2^{b}-1}\cdot3\sigma| (x\sim N(0,\sigma^2),  x>0)$.

Thus, for $x\sim N(0,\sigma^2)$, we get "\emph{the quantization error would increase by the possibility of $P(x>\frac{2^{b'}-1}{2^{b}-1}\cdot3\sigma| (x\sim 2N(0,\sigma^2),  x>0)$}". \textbf{Theorem 1} holds. So does \textbf{Lemma 1}.

\subsection{Gradient calculations of $V_{th,l}^1$ and $S_q^l$}
\label{sec:step gradient}
We adopt the same gradient calculations of the quantization step size of \citep{esser2020learned}.

For  $V^1_{th,l}$,
\begin{align}
&\frac{\partial L_{task}}{\partial V^1_{th,l}} =  \sum^T_t \sum_j \frac{1}{V^1_{th,l}} \frac{\partial L_{task}}{\partial S^t_{out,l}}  \cdot g_{s,scale}  \cdot g_{q}, \\ 
&g_{q} =
\begin{cases}
\left\lfloor\frac{v^t_l}{V^1_{th,l}}+V_{th,l}^2\right\rfloor - \frac{v^t_l}{V^1_{th,l}}, \frac{v^t_l}{V^1_{th,l}} \in [q_{min},q_{max}]\\
0, \frac{v^t_l}{V^1_{th,l}}<q_{min}\\
q_{max}, q_{max}<\frac{v^t_l}{V^1_{th,l}},
\end{cases}
\end{align}
where the minimum quantization integer $q_{min}=0$ and the maximum integer $q_{max}=2^{B_{s,l}}-1$. While, $g_{s,scale}$, equal to $\frac{1}{\sqrt{\sum_j1\cdot q_{max}}}$ and meant to avoid gradient explosion \citep{esser2020learned}, is the gradient scaling factor. $L_{task}$ represents the task loss signal.

For  $S^l_q$,
\begin{align}
&\frac{\partial L_{task}}{\partial S_q^l} =  \sum_i \frac{\partial L_{task}}{\partial w^l_q} \cdot g_{w,scale} \cdot g_{q}, \\ 
& g_{q}= 
\begin{cases}
\left \lfloor \frac{w_l}{S^l_q} \right\rceil - \frac{w_l}{S^l_q}, \frac{w_l}{S^l_q} \in [q_{min}, q_{max}]\text{ and }B_{w,l}>1\\
sign(\frac{w_l}{S^l_q}) - \frac{w_l}{S^l_q}, \frac{w_l}{S^l_q}\in [q_{min}, q_{max}]\text{ and }B_{w,l}=1\\
q_{min}, \frac{w_l}{S^l_q}<q_{min}\\
q_{max}, q_{max}<\frac{w_l}{S^l_q}.
\end{cases}
\end{align}
Specially, when $B_{w,l}=1$, $q_{min}=-1$ and $q_{max}=1$. Otherwise, $q_{min}=-2^{B_{w,l}-1}+1$ and $q_{max}=2^{B_{w,l}-1}-1$. Similarly,  $i$ is the index of the weight element, and $g_{w,scale} = \frac{1}{\sqrt{\sum_i1\cdot q_{max}}}$.

\subsection{Gradient calculations of $B_{s,l}$ and $V_{th,l}^1$ of the bidirectional neuron}
\label{sec:2 bi gradient}
Since Equation~\ref{eq:our bsn3} has a slightly different formulation, the corresponding gradient calculations of $B_{s,l}$ and $V_{th,l}^1$ would also change a bit. 

For $B_{s,j}$,
\begin{align}
    &\frac{\partial L_{task}}{\partial \hat{B}_{s,l}} = \frac{\partial L_{task}}{\partial B_{s,l}}=  \sum_j \frac{\partial L_{task}}{\partial S_{put,l}^t}  \cdot g_{s,scale} \cdot g_{q}, \\ 
&g_{q} =
\begin{cases}
sign( \frac{v^t_l}{V_{th,l}^1}) \cdot (q_{max}+1)   \cdot    \ln2   , \frac{v^t_l}{V_{th,l}^1}\notin[q_{min}, q_{max}] \\0, \text{otherwise}.
\end{cases}
\end{align}
Here,  $j$ is the index of the weight element, and $g_{s,scale} = \frac{1}{\sqrt{\sum_j1\cdot q_{max}}}$.
Specially, when $B_{s,l}=1$, $q_{min}=-1$ and $q_{max}=1$. Otherwise, $q_{min}=-2^{B_{s,l}-1}+1$ and $q_{max}=2^{B_{s,l}-1}-1$.

For $V_{th,l}^1$,
\begin{align}
&\frac{\partial L_{task}}{\partial V_{th,l}^1} =  \sum_j \frac{\partial L_{task}}{\partial S^t_{out,l}} \cdot g_{s,scale} \cdot g_{q}, \\ 
& g_{q}= 
\begin{cases}
\left \lfloor \frac{v_l^t}{V_{th,l}^1} \right\rceil - \frac{v_l^t}{V_{th,l}^1}, \frac{v_l^t}{V_{th,l}^1} \in [q_{min}, q_{max}]\text{ and }B_{s,l}>1\\
sign(\frac{v_l^t}{V_{th,l}^1}) - \frac{v_l^t}{V_{th,l}^1}, \frac{v_l^t}{V_{th,l}^1}\in [q_{min}, q_{max}]\text{ and }B_{s,l}=1\\
q_{min}, \frac{v_l^t}{V_{th,l}^1}<q_{min}\\
q_{max}, q_{max}<\frac{v_l^t}{V_{th,l}^1}.
\end{cases}
\end{align}
Here, $j$ is the index of the feature map element, $q_{min}=-2^{B_{s,l}-1}+1$, and $q_{max}=2^{B_{s,l}-1}-1$. $g_{scale}=\frac{1}{\sqrt{\sum_j1\cdot q_{max}}}$.

\subsection{Details of the experimental settings}
\label{sec: exp set}
All our experiments are conducted on A100 GPUs. 

\subsubsection{CIFAR}
In the CIFAR experiments, two models are adopted ResNet20 and Spikformer.  For ResNet20, we adhere to the default settings of \citep{guo2024ternary} and \citep{guo2022real}. The training epochs are set to 200, batch size to 128, and learning rate to 0.1. Besides, the penalty coefficients $\lambda_1$, $\lambda_2$, and $\lambda_3$ are set to 4e-2, 4e-2, and 1e-2, respectively. The initial W/S/T are set to 4/4/2. These settings apply to both comparative and ablation studies. For Spikformer, we adopt the same default settings  as \citep{zhou2022spikformer} from their official codes, including every training configuration. The penalty coefficients and the initial W/S/T are the same as our ResNet setting.

Additionally, $B_{w,bound}$, $B_{s,bound}$, and $T_{bound}$ are set to 6,6, and 3, respectively. 

\subsubsection{ImageNet}
In the ImageNet-1k experiments, we exclusively adopt the SEWResNet34 model. The training codes are also built on the default \citep{guo2024ternary}. The training epochs are set to 200, batch size to 128$\times$4 (data distributed parallel on 4 GPU cards), and learning rate to 0.1. Besides, the penalty coefficients $\lambda_1$, $\lambda_2$, and $\lambda_3$ are set to 8e-2, 8e-2, and 1e-2, respectively. Following those advanced SNNs, e.g., \citep{shen2024conventional} and \citep{yao2023attention}, we directly set time-step=1 to demonstrate our model's superior accuracy on large-scaled ImageNet, i.e., the initial W/S/T are set to 4/4/1. 
In Section~\ref{sec:add exps}, we also show the result of the initial W/S/T=4/4/2.

Additionally, $B_{w,bound}$, $B_{s,bound}$, and $T_{bound}$ are set to 6,6, and 2, respectively.

\subsubsection{CIFAR10-DVS}

In the CIFAR10-DVS experiments, we only use the default DVS Spikformer defined by the official codes of \citep{zhou2022spikformer}. To better observe the temporal dimension's influence on the model accuracy of the event dataset. We set $T_l$ back to the unlearnable constant and the corresponding temporal squeezing is also deactivated.
The corresponding penalty coefficients $\lambda_1$ and $\lambda_3$ are all set to 4e-2. The target weight bit widths are all set to 1, the target spike bit widths are 2 or 4. Under this setting, SNNs can fully exploit the model-level sequential processing ability. Thus, we can accurately observe that our methods can mitigate "the temporal constraint" steadily.
 In Section~\ref{sec:add exps}, we also list the results of applying the learnable temporal lengths and temporal squeezing.

Additionally, $B_{w,bound}$ and $B_{s,bound}$ are both set to 6. And, W/S are initial to 4/4. For experiments with learnable $T_l$ in Table~\ref{exp:more exp on dvs}, we cancel the temporal-wise sharing and reset $\lambda_1$ to 8e-2.

\subsubsection{DVS-GESTURE}

In the DVS-GESTURE experiments, we also use the default DVS Spikformer defined by the official codes of \citep{zhou2022spikformer}. Most experimental settings are the same as the CIFAR10-DVS experiments and the official codes \citep{zhou2022spikformer}. Following is some exceptions: the training epoch is set to 192, and when the target time length is 1, 2, or x (x>2), each event data is divided into 4, 10, or 16 temporal slices, respectively. Only the first 1, 2, or x temporal slices are extracted to training and inference.

\subsubsection{Additional statements about the quantization baselines}
The closest baseline-work to ours is \citep{shen2024conventional}. However, we find that their work is not open-sourced and lacks a lot of implementation details. Therefore, we try to reproduce their method based on their paper and our practice as we call "U-quant." in the main text. Nevertheless, we also find many of their statistics, such as S-ACE, may be roughly calculated so we have to re-calculate every statistics we show in the main text with more accurate means. 

In addition, we also stop citing some of their results that we fail reproducing, and report the "U-quant." results instead.

\subsection{Elaborations on how temporal squeezing would save S-ACE}
\label{sec:elab on ts}

\begin{figure}[t]
  \centering
  \includegraphics[width= .9\linewidth]{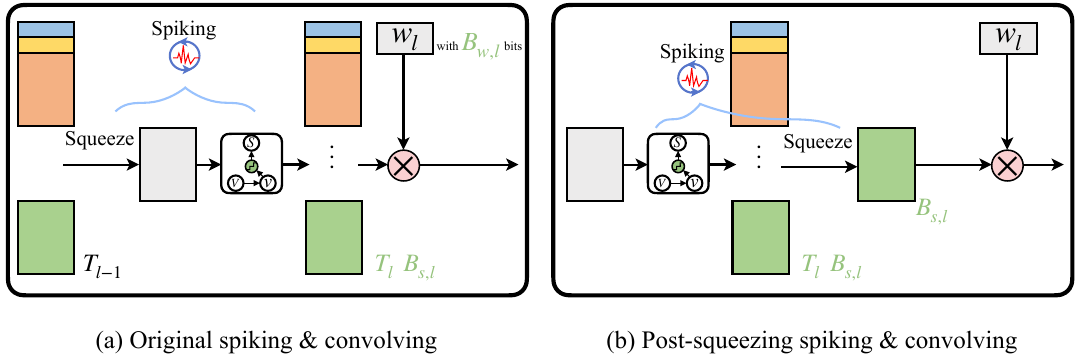}
  \caption{Different computation pipelines with the temporal squeezing. \textbf{(a)} the conventional computing pipeline. The input current is squeezed before the spiking operation. \textbf{(b)} the identical post-squeezing pipeline. The temporal squeezing operation is conducted after the spiking operation.}
  \label{fig:squeeze}
\end{figure}
Our proposed temporal squeezing uses the mean operation to extract the temporal information.

One merit of using the mean operation is the squeezed data will not require any extra bit width. For instance, for a positive spike train $S_{out}=\{S^1_{out}, S^2_{out},...,S^t_{out},...,S^T_{out}\}$. Each spike $S^t_{out}$ has $B_{s}$ bits. To contain the squeezed spike map $\hat{S}_{out}=\frac{1}{T}\sum^T_{t=1} S^t_{out}$ without any loss,  $B_{s}$ bits would be enough. This gives us the opportunity to reduce the S-ACE of SNNs' convolution by changing the computing pipeline.

As shown in Figure~\ref{fig:squeeze}, the conventional spiking-and-convolving pipeline will require $T_l\times B_s \times B_w \times MAC$ S-ACEs.  With the changed post-squeezing pipeline, 
we only need $ B_s \times B_w \times MAC$ S-ACEs to implement the identical computation. Thus, when the overall $T_l$'s of SNNs are over 2, such computation advantage of the temporal squeezing would be more pronounced.

\subsection{Additional experimental results on ImageNet and CIFAR10-DVS}
\label{sec:add exps}

In this section, we add more experimental results as expanded evidence to validate more aspects of our methods.

Firstly, as described in Section~\ref{sec: exp set}, we use the initial T=1 in the main text. This may raise the concern about whether the temporal squeezing cannot function well on such up-scaled dataset. As shown in Table~\ref{exp:more exp on imagenet}, with the temporal squeezing, the model can achieve the same model accuracy while using fewer S-ACEs, handling the change of temporal length from 2 to 1 during training well. As for the reason why the model initialized as W/S/T=4/4/2 can perform better on accuracy very slightly. One explanation is that
the initial temporal length does not affect the model performance while the target temporal length (or the actual temporal length) does, which is in line with the prior finding of \citep{shen2024conventional} that the temporal length rarely affects the performance of multi-bit SNNs on static datasets. Therefore, for the saving of training time, we choose to set the initial temporal length as 1 in the experiments of the main text since the longer initial temporal length causes longer forward and backward inference time. Additionally, from another perspective, this could further prove that our multi-bit SNN has advanced its spatial processing ability to the limit, one time-step would suffice to fulfill the requirement of a high-accuracy inference. 

In addition, we also naively apply our methods to Spikformer-8-512 on ImageNet with the same model hyperparameters as \citep{shen2024conventional} and \citep{zhou2022spikformer}. Their models achieve 70.87\% \citep{shen2024conventional} and 73.38\% \citep{zhou2022spikformer}  with W/S/T = 4/2/1 and  16/1/4, respectively. While, our Spikformer achieves a higher 74.87\% top-1 accuracy with lower bit budgets that W/S/T = 2.64/2.21/1.0.

\begin{table}[width=\linewidth,cols=6,pos=h]
\centering
\caption{Additional experimental results on ImageNet.}
\label{exp:more exp on imagenet}
\begin{adjustbox}{max width=\linewidth}
\begin{threeparttable}
\begin{tabular}{cccccc}
\toprule
\textbf{Method}           & \textbf{Architecture}         & \textbf{W/S/T}         & \textbf{Bit Budget} & \textbf{S-ACE} & \textbf{Top-1} \\ \midrule
Attention-SNN             & ResNet34                      & 16/1/1                 & 16                  & 56.73          & 69.15          \\
TET                       & Sew-ResNet34                  & 16/1/4                 & 64                  & 226.93         & 68.00          \\
Ternary Spike             & ResNet34                      & 16/2/4                 & 128                 & 435.86         & 70.74          \\
\multirow{2}{*}{PLIF+SEW} & SEW-ResNet34                  & 16/1/4                 & 64                  & 226.93         & 67.04          \\
                          & SEW-ResNet50                  & 16/1/4                 & 64                  & 215.62         & 67.78          \\ \midrule
Quant. SNN                & SEW-ResNet34                  & 8/8/1                  & 64                  & 226.93         & 70.13          \\ \midrule
Ours-441-221              & \multirow{3}{*}{SEW-ResNet34} & \textbf{2.58/2.18/1.0} & \textbf{5.61}       & \textbf{30.74} & 70.57          \\
Ours-441-241              &                               & \textbf{2.62/3.96/1.0} & \textbf{10.39}      & \textbf{54.69} & \textbf{72.02} \\
Ours-442-241              &                               & \textbf{2.60/4.01/1.0} & \textbf{10.41}      & \textbf{54.26} & \textbf{72.11} \\ \bottomrule
\end{tabular}
\begin{tablenotes}
\footnotesize
\item "Ours-xxx-yyy" denotes the initial W/S/T=x/x/x and the target W/S/T=y/y/y.  
\end{tablenotes}
\end{threeparttable}
\end{adjustbox}
\end{table}

Secondly, as stated in Section~\ref{sec: exp set}, we use the conventional sequential-processing pipeline to address the event dataset in the main text, i.e., the temporal squeezing and learnable $T$ are canceled. In this way, we can accurately control the temporal length and fully utilize the model-level temporal processing of SNN. This may raise the concern that the temporal squeezing could severely impair the temporal processing ability. As shown in Table~\ref{exp:more exp on dvs}, we find the temporal squeezing can still outperform the quantization baseline \citep{shen2024conventional} and maintain good performance under low time-steps, although the overall model performance, compared to  the case without temporal squeezing,  is decreased. One possible explanation is that although the model-level temporal processing ability may be diminished, the layer-wise  temporal processing ability can still be maintained when using temporal squeezing. And, temporal squeezing also meets the temporal information dynamics that as the training goes on, more valuable and decisive information aggregates into the early time-steps \citep{Kim_Li_Park_Venkatesha_Hambitzer_Panda_2023}. As listed in Table~\ref{exp:more exp on dvs}, the best results of temporal squeezing show up under the low time-step settings. Temporal squeezing merely accelerates the aggregation process. That is why  temporal squeezing would not severely undermine the sequential-processing ability and perform well even on the sequential dataset.

\begin{table}[width=\linewidth,cols=8,pos=h]
\centering
\caption{Additional experimental results of Spikformer (Spf.) on CIFAR10-DVS.}
\label{exp:more exp on dvs}
\begin{adjustbox}{max width=\linewidth}
\begin{threeparttable}
\begin{tabular}{cccccccc}
\toprule
\multicolumn{2}{c}{\textbf{Origin}\citep{shen2024conventional}}             & \multicolumn{2}{c}{\textbf{Quant. Spf.}\citep{shen2024conventional}} & \multicolumn{2}{c}{\textbf{Our Spf.}} & \multicolumn{2}{c}{\textbf{Our Spf. w/ Learnable $T_l$}} \\ \midrule
\textbf{W/S/T}           & \textbf{Top-1}        & \textbf{W/S/T}      & \textbf{Top-1}      & \textbf{W/S/T}    & \textbf{Top-1}    & \textbf{W/S/T}            & \textbf{Top-1}            \\ \midrule
\multirow{5}{*}{16/1/16} & \multirow{5}{*}{80.7} & 1/1/16              & \textbf{79.8}       & 1.11/2.00/6.0     & \textbf{80.9}     & 1.14/1.97/6.00            & 78.7                      \\
                         &                       & 1/2/8              & 79.3                & 1.28/2.00/4.0     & 80.0              & 1.14/1.97/4.00            & 79.3                      \\
                         &                       & 1/4/4               & 63.1                & 1.33/4.00/3.0     & 80.1              & 1.17/4.08/2.05            & 79.6                      \\
                         &                       & 1/8/2               & 43.0                & 1.31/4.03/2.0     & 78.7              & 1.09/1.95/1.95            & \textbf{80.0}             \\
                         &                       & 1/16/1              & 35.8                & 1.31/4.03/1.0     & 77.4              & 1.11/1.49/1.00            & 77.3                      \\ \bottomrule
\end{tabular}

% \begin{tablenotes}
% \footnotesize
% \item $\dagger$ denotes plain uniform quantization, following  \citep{shen2024conventional}.
% \item $*$ denotes the temporal squeezing is applied in inference.
% \end{tablenotes}
\end{threeparttable}
\end{adjustbox}
\end{table}

\textbf{Additional statements on temporal squeezing.} Although we have clarified the purpose of using temporal squeezing in the main text, we would like to highlight it here one more time.
The idea of proposing temporal squeezing is tackling the interlayer data mismatching issue and deriving the gradient for $T_l$. Only in this way could the temporal bit widths be adaptively allocated throughout different layers and make dataflow smooth. Temporal squeezing is more of a technical resolution for applying the mixed-precision framework to SNNs. We do not expect this technique can bring any model performance advancement

Our main contributions are refining the spiking neuron, the formulation of the temporal-related step-size mismatch issue, and the step-size renewal mechanism. All of them are validated through various experiments as we showed in the main text.

{

\subsection{Potential exploration to spiking audio task} \label{sbsec:shd sec}

We also conduct a simple experiment on the spiking audio dataset SHD \citep{cramer2020heidelberg}.
Specifically, we plug our method to  the implementation of dilated convolution with learnable spacings (DCLS) \citep{hammouamri2023learning}, which is a synaptic-delay based framework to help SNNs operate on spiking audio tasks. 

As shown in Table~\ref{exp:exp on shd}, our method can help the original DCLS baseline to achieve better model accuracy, demonstrating a potential compatibility of our method with spiking audio task.

Additionally, we follow the exactly same hyperparameters and model architectures of DCLS \citep{hammouamri2023learning} without changing a thing. The corresponding penalty coefficient $\lambda_1$ is set to 4e-2. The initial W is set to 6 and target Ws are set to 3 and 4 in Table~\ref{exp:exp on shd}. The initial and target Ss are all set to 1.  The initial and target Ts are all set to 140. 

\begin{table}[width=\linewidth,cols=5,pos=t]

\centering
\caption{{Extended experiments on the spiking audio dataset SHD. }}
\label{exp:exp on shd}
\begin{adjustbox}{max width=\linewidth}
{
\begin{threeparttable}\begin{tabular}{ccc}
\toprule
\textbf{Method} & \textbf{Parameters (M)} & \textbf{Top-1}  \\ \midrule
DMUC \citep{sun2024delayed}       & 0.24                    & 91.48          \\
CNN \citep{cramer2020heidelberg}            & 1.01                    & 92.40          \\
RSNN \citep{cramer2020heidelberg}            & 1.97                  & 83.20          \\
Attention RSNN \citep{yao2021temporal}         & 0.14                     & 91.08          \\
RadLIF \citep{bittar2022surrogate}         & 3.9                     & 94.62          \\
DL256-SNN-Dloss \citep{sun2023learnable} & 0.21                    & 93.55          \\ \midrule
DCLS (T=140) \citep{hammouamri2023learning}         & 0.21                    & 95.07          \\ \midrule
Our DCLS-3.5/1/140   & 0.21           & 95.17 \\
Our DCLS-4.4/1/140   & 0.21           & \textbf{95.54} \\ \bottomrule
\end{tabular}
\begin{tablenotes}
\footnotesize
\item "Our DCLS-x/x/x" denotes the learned bit widths of W/S/T.
% \item Following the research tradition of spiking audio tasks, we only report parameter number (M) instead of model size (MB).
\end{tablenotes}
\end{threeparttable}}
\end{adjustbox}
\end{table}
}

\subsection{Maintaining the asynchronous nature of SNNs}
\label{sec:maintain asyn}
The most precious characteristic that SNN brings is its asynchronous processing ability, i.e., event-driven and clock-free computing. We would like to address the possible concerns about how such nature could be maintained when temporal squeezing is applied and could there exist potential issues.

\textbf{Exploiting the spike perceptrons.}  \emph{Temporal squeezing itself is consistent with neurodynamics as it is essentially composed of a signal accumulation \citep{gerstner2014neuronal} and a division operation.} A similar accumulation mechanism can also be found in the current accumulation stage of spiking neuron, such as LIF and IF. Particularly, SpikeConverter neuron \citep{liu2022spikeconverter} separates this accumulation from the spiking neuron modeling and uses a particular computing pipeline to handle this accumulation. Their accumulation is identical to ours, which is proven feasible in asynchronous computing.
In addition, such accumulation has also been realized and implemented by neuromorphic circuits to process the DVS signals and generate the dynamic datasets \citep{amir2017low,lin2021imagenet}. 

As such, we can simply use the same signal accumulation diagrams, i.e.  the spike perceptrons, to realize temporal squeezing as what has been used
in DVS cameras \citep{li2017cifar10} and previous spiking neurons \citep{liu2022spikeconverter}. In other words, such different computing paradigm as shown in Figure~\ref{fig:squeeze} can be implemented under the existing neuromorphic computing and will not cause any loss of the asynchronous and clock-free nature.

\begin{algorithm}[h]
	% \textsl{}\setstretch{1.8}
	\renewcommand{\algorithmicrequire}{\textbf{Input:}}
	\renewcommand{\algorithmicensure}{\textbf{Output:}}
    \caption{Overall training pipeline of the proposed adaptive-bit-allocation SNNs.}
    \label{algo:overall}
    \begin{algorithmic}[1]
		\REQUIRE Target W/S/T $B^{tar}_w$, $ B^{tar}_s$, and $T^{tar}$; penalty coefficients $\lambda_1$, $\lambda_2$, and $\lambda_3$; full-precision SNN with our refined spiking neurons, $L$ layers, and weights $w_l$; the actual bit widths of each layer $B_{s,l}^t$, $B_{w,l}$, and $T_l$; the corresponding quantization step sizes $V_{th,l}^{1,t}$ and $S_q^l$; the training dataset $D$; the training epochs $ep$; the training batch size $bs$.
  
		\ENSURE mixed-precision SNN.
            \STATE Initialize  the bit-width observers in Algorithm 1.
            \STATE Initialize  $V_{th,l}^{1,t}$ and $S_q^l$ using the first mini-batch of $D$.
            \STATE The shutting thresholds $Th_w \gets 0.24|\bar{B}_{w,l}-B^{tar}_w| $ and  $Th_s \gets 0.24|\bar{B}_{s,l}^t-B^{tar}_s| $.
            \STATE Renewal flags $f_w \gets 0$ and $f_s \gets 0$
            % \STATE Recording the initial bit  $B^{'t}_{s,l}\gets B_{s,l^t}$ and $B'_{w,l}\gets B_{w,l}$.
            \STATE for $i=1; i\le ep; i++$ do
            \STATE \quad for $j=1; j\le ceil(\frac{D}{bs}); j++$ do
            \STATE \quad \quad for $l=1; l\le L; l++$ do
            \STATE \quad \quad  \quad    \# the step-size renewal of weight
            \STATE \quad \quad  \quad   if $|(\bar{B}_w-B^{tar}_w| > Th_w$ and $f_w == 0$ do
            % \STATE \quad \quad  \quad \quad \# the step-size mechanism  (Algorithm 1)
            % \STATE \quad \quad  \quad \quad if $B_{w,l}\ne B'_{w,l}$ do
            % \STATE \quad \quad  \quad \quad \quad $B'_{w,l}\gets B_{w,l}$
            \STATE \quad \quad  \quad \quad  $S_q^l\gets$ RA($B_{w,l}$, $w_l$)
            \STATE \quad \quad  \quad   else do
            \STATE \quad \quad  \quad \quad $f_w \gets 1$
            \STATE \quad \quad  \quad    \# the step-size renewal of spike activation
            \STATE \quad \quad  \quad   if $|(\bar{B}_s-B^{tar}_s| > Th_s$ and $f_s == 0$ do
            % \STATE \quad \quad  \quad \quad \# the step-size mechanism  (Algorithm 1)
            % \STATE \quad \quad  \quad \quad if $B_{s,l}^t\ne B_{s,l}^{'t}$ do
            % \STATE \quad \quad  \quad \quad \quad $B_{s,l}^{'t}\gets B_{s,l^t}$
            \STATE \quad \quad  \quad \quad  $V_{th,l}^{1,t}\gets$ RA($B_{s,l}^t$, $v_l^t$)
            \STATE \quad \quad  \quad   else do
            \STATE \quad \quad  \quad \quad $f_s \gets 1$
            \STATE \quad \quad  \quad    $S_{out,l}\gets$  SNN$_l$($S_{out,l-1},V_{th,l}^{1,t},w_l,S_q^l$)
            \STATE \quad \quad   $L_{total}\gets L_{task}+\lambda_1L(\bar{B}_w, B^{tar}_w)+\lambda_2L(\bar{T}, T^{tar})+\lambda_3L(\bar{B}_s, B^{tar}_s)$
            \STATE \quad \quad  Back-propagation and updating SNN parameters
    \end{algorithmic} RA denotes the renewal algorithm (Algorithm 1). $\bar{x}$ denotes the averaged value of $x$ over the whole net.
\end{algorithm}

\textbf{Using the two-stage pipeline or keeping single time-step.} Although we have theoretically elaborated on the harmlessness of temporal squeezing to the asynchronous nature in the above, one may argue that the entire network's spike flow could be interrupted by the squeezing stage, leading to longer inference latency. This concern has been discussed and resolved by \citep{liu2022spikeconverter}, using the two-stage pipeline. The first stage is spike accumulation, i.e., the temporal squeezing, the second stage is spike emission. These two separate stages can overlapped inter-layer. Specifically, assuming spike accumulation and emission both require 4 hardware clock cycles for computing. After the first cycle of the spike emission of the previous layer, the first computing cycle of the spike accumulation of the next layer can be started instead of waiting for the whole 4 cycles of the spike emission of the previous layer being finished. Thus, 3 cycles are overlapped, and the actual latency of the two-stage spiking pipeline is only 5 cycles, only one more cycle that the ordinary spiking pipeline. Therefore, using temporal squeezing only introduces very limited latency increase. 

Such concern can also be resolved by the fact that we only use one or two time-steps to achieve the most advanced model accuracy on static datasets as shown in the main text. As for the event dataset, we have also mitigated the temporal constraint. High accuracy can be maintained for the event data, even when the temporal lengths are extremely short. Therefore, the model-level asynchronous dataflow can be well maintained in our framework without drastically longer-latency issue.

\textbf{Canceling temporal squeezing.} Even in the worst-case assumption that temporal squeezing had no neuromorphic hardware support and must be applied to the long temporal length, we could simply cancel this mechanism and give up the learnable $T_l$ as the solution because as we have stated before:  temporal squeezing is more of a technical resolution for applying the mixed-precision framework to SNNs. Moreover, our main contributions are refining the spiking neuron, the formulation of the temporally related step-size mismatch issue, and the step-size renewal mechanism. As the experiments of Table~\ref{exp:more exp on dvs} on the event dataset show our methods can achieve better model performance without temporal squeezing, and temporal squeezing, aiming at realizing learnable and layer-wise different $T_l$, will not undermine the temporal-processing ability of SNNs as well.

\subsection{Overall algorithm of the proposed bit allocation}
\label{sec:overall algo}
For clarity, we provide the pseudo code of the overall algorithm, as shown in Algorithm~\ref{algo:overall}.

\printcredits

\bibliographystyle{cas-model2-names}

\bibliography{cas-refs_v2}

% v.s. kip3pt

% Author biography without author photo.
% Author biography. Author biography. Author biography.
% Author biography. Author biography. Author biography.
% Author biography. Author biography. Author biography.
% Author biography. Author biography. Author biography.
% Author biography. Author biography. Author biography.
% Author biography. Author biography. Author biography.
% Author biography. Author biography. Author biography.
% Author biography. Author biography. Author biography.
% Author biography. Author biography. Author biography.

% \bio{figs/cas-pic1}
% Author biography with author photo.
% Author biography. Author biography. Author biography.
% Author biography. Author biography. Author biography.
% Author biography. Author biography. Author biography.
% Author biography. Author biography. Author biography.
% Author biography. Author biography. Author biography.
% Author biography. Author biography. Author biography.
% Author biography. Author biography. Author biography.
% Author biography. Author biography. Author biography.
% Author biography. Author biography. Author biography.

% \bio{figs/cas-pic1}
% Author biography with author photo.
% Author biography. Author biography. Author biography.
% Author biography. Author biography. Author biography.
% Author biography. Author biography. Author biography.
% Author biography. Author biography. Author biography.

\end{document}